%% file: main.tex
\documentclass[10pt,twocolumn,letterpaper]{article}

\usepackage[pagenumbers]{cvpr} %
\usepackage[accsupp]{axessibility}  %

\input{preamble}
\definecolor{cvprblue}{rgb}{0.21,0.49,0.74}
\usepackage[pagebackref,breaklinks,colorlinks,allcolors=cvprblue]{hyperref}

\usepackage{url}
\usepackage{tabularx}
\usepackage{multirow}
\usepackage{pifont}%
\usepackage{microtype}
\usepackage{xcolor,colortbl}

\def\paperID{16033}
\def\confName{CVPR}
\def\confYear{2026}

\title{Investigating self-supervised representations for audio-visual deepfake detection}

\author{
    Dragos-Alexandru Boldisor\footnotemark[1] $^{1,2}$ ~~~
    Stefan Smeu\thanks{Equal contribution.} $^{1}$ ~~~
    Dan Oneata$^{2,1}$ ~~~
    Elisabeta Oneata$^1$ \\
    $^1$Bitdefender ~~~~~
    $^2$\textsc{Politehnica} Bucharest \\
    {\tt\small\{ssmeu, dboldisor, eoneata\}@bitdefender.com} ~~
    {\tt\small dan.oneata@gmail.com}
}

\newcommand{\mypar}[1]{\vspace{2mm}\noindent\textbf{#1}}

\newcommand{\cmark}{\ding{51}}

\definecolor{my_green}{HTML}{339933}
\definecolor{my_red}{HTML}{eb9898}
\definecolor{my_green2}{HTML}{e1f5e6}

\newcommand{\wb}{\mathbf{w}}
\newcommand{\xb}{\mathbf{x}}
\newcommand{\rva}{\mathbf{a}}
\newcommand{\rvv}{\mathbf{v}}

\newcommand{\na}{\color{gray}\textsc{n/a}}
\newcommand{\gray}[1]{\color{gray}#1}
\newcommand{\ii}[1]{{\scriptsize{\gray \sf #1}}}
\newcommand{\mylabel}[1]{{\small \gray \sf #1}}

\begin{document}
\maketitle

\begin{abstract}
    Self-supervised representations excel at many vision and speech tasks, but their potential for audio-visual deepfake detection remains underexplored.
    Unlike prior work that uses these features in isolation or buried within complex architectures, we systematically evaluate them across modalities (audio, video, multimodal) and domains (lip movements, generic visual content).
    We assess three key dimensions: detection effectiveness, interpretability of encoded information, and cross-modal complementarity.
    We find that most self-supervised features capture deepfake-relevant information, and that this information is complementary. %
    Moreover, models primarily attend to semantically meaningful regions rather than spurious artifacts (such as the leading silence).
    Among the investigated features, audio-informed representations generalize best and achieve state-of-the-art results.
    However, generalization to realistic in-the-wild data remains challenging.
    Our analysis indicates this gap stems from intrinsic dataset difficulty rather than from features latching onto superficial patterns.
    Project webpage: {\small \url{https://bit-ml.github.io/ssr-dfd}}.
\end{abstract}

\section{Introduction}
\label{sec:introduction}

Generative models now produce text, images, audio, video, all rivaling human creations.
This progress brings a new challenge: detect whether content is authentic or machine-generated (deepfake).
Reliable detection prevents obvious risks such as disinformation and fraud, but also serves a simple need:
users want to know what they can trust.
We tackle the detection problem in the video domain, the internet's most consumed medium
and, in particular, we focus on people speaking, the most consequential form of deepfakes.

\begin{figure}
    \centering
    \includegraphics[width=\linewidth]{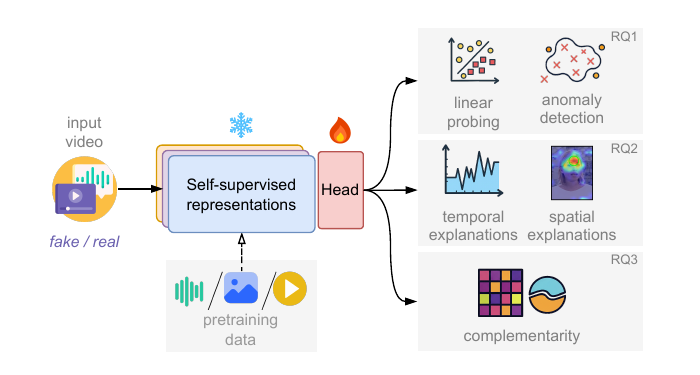}
    \caption{%
        We evaluate a wide array of self-supervised representations for audio-visual deepfake detection.
        We design a multi-faceted evaluation to understand their:
        (i) usefulness and robustness (via linear probing and anomaly detection),
        (ii) focus and interpretability (via temporal and spatial explanations), and
        (iii) complementarity (via correlation and fusion analyses).
    }
    \label{fig:teaser}
\end{figure}

Many approaches are being continuously proposed for the task of audio-visual deepfake detection: these range
from powerful discriminative classifiers \cite{koutlis2024dimodiff,nie2024frade,wu2025hola,10743671}
to techniques that exploit inconsistencies between modalities \cite{feng23cvpr,reiss2023detectingdeepfakesseeing,liang24speechforensics,li2024zero}.
Yet, the backbone features underlying these models often determine their effectiveness.
Recent work shows strong results of self-supervised learning in this context:
image-based detectors benefit from CLIP \cite{ojha2023towards,cozzolino2024cvprw},
audio-based detectors from Wav2Vec2 \cite{wang22odyssey,pascu24interspeech}, and
audio-visual models from AV-HuBERT \cite{reiss2023detecting,liang24speechforensics}. %
The self-supervised representations capture rich, modality-specific structure without requiring task-specific supervision, making them especially attractive for deepfake detection.

In this paper, we evaluate for the first time a wide range of self-supervised features (audio-only, image-only, multimodal) for the task of audio-visual deepfake detection.
Our aim is to understand what these features capture and how they contribute to detection performance.
We center our work around three research questions:
\begin{enumerate}
    \item \textbf{How useful are self-supervised features for deepfake detection?}
        Do they %
        generalize across domains and to the related task of anomaly detection?
    \item \textbf{Where do self-supervised features look?}
        Does the model attend to the manipulated regions?
        Does it align with human annotations?
    \item \textbf{How complementary are different features?}
        If multiple feature types succeed at detection, do they rely on similar cues or do they encode distinct information?
\end{enumerate}

To answer these questions, we start by adapting linear probing \cite{alain2017,hupkes2018,belinkov2022a} to the video domain.
By keeping the classifier minimal, we can directly measure the information already encoded in the feature representations.
We evaluate both on standard scientific datasets \cite{khalid2021fakeavceleb,cai2024avdeepfake1mlargescalellmdrivenaudiovisual,liu2024lips},
but also on challenging in-the-wild data \cite{chandra2025deepfake},
assessing both in-domain and out-of-domain performance.

However, prior work has warned that even subtle distribution shifts between real and fake samples can be exploited by classifiers, leading to spurious correlations \cite{chai2020eccv,muller2022does,Smeu_2025_CVPR,klemt2025deepfake}.
Worse still, such spurious cues may persist across datasets, yielding overly optimistic results.
We therefore propose a multi-faceted evaluation suite (\cref{fig:teaser}) that goes beyond standard supervised testing and more directly tests whether the representations capture meaningful forensic cues.

First, we assess feature quality via the proxy task of anomaly detection.
By training only on real samples, we break the real–fake asymmetry and avoid relying on spurious cues.
We consider two approaches:
(i) density estimation via next-token prediction, and
(ii) audio–video synchronization, which measures cross-modal alignment between representations.
Second, to further probe model behavior, we extract implicit localizations from the classifier.
We apply temporal and spatial explainability methods and test whether the resulting explanations align
with temporal manipulations \cite{cai2024avdeepfake1mlargescalellmdrivenaudiovisual} and human annotations \cite{hondru2025exddv}, respectively.
Finally, we examine whether the representations capture different information by measuring the correlation between their predictions and the downstream performance of their combination.

Our main findings are as follows:
1. Many of the investigated features obtain strong in-domain detection performance.
Audio-informed representations generalize best and achieve state-of-the-art results.
2. Training only on real data helps identify features that rely solely on spurious cues,
but remain less effective for deepfake detection than supervised learning.
3. Temporal explanations indicate that many features capture meaningful cues, even in the presence of spurious correlations.
Spatial explanations are plausible and partially match human annotations.
4. Visual representations are more complementary than audio ones.
Combining representations yields gains that tend to increase with complementarity.

\section {Related work}
\label{sec:related_work}

\mypar{Self-supervised learning} (SSL) learns powerful representations by solving pretext tasks on large-scale unlabeled data
\cite{balestriero2023cookbook}.
These representations transfer effectively to many downstream tasks, such as classification or anomaly detection.
In the \textbf{visual} domain \cite{uelwer2025survey}, many approaches use contrastive learning %
between
images and their augmented versions \cite{he2020moco, chen2020simple} or images and their captions \cite{jia21align, radford2021learning}.
Alternatives include masked image modeling \cite{he2022masked} or discriminative self-distillation \cite{oquab23dinov2,simeoni2025dinov3}.
For video, spatio-temporal information is used to build SSL representations \cite{qian2021spatiotemporal, tong2022videomae}.
In the \textbf{audio} domain \cite{liu2022audio}, Wav2Vec \cite{baevski2020wav2vec,babu22xlsr2} and HuBERT \cite{hsu21hubert} introduced predictive coding and masked prediction for speech signals, significantly improving ASR with limited labeled data.
For \textbf{audio-visual} data,
AV-HuBERT \cite{shi2022learning} extends HuBERT to learn joint speech--lip representations. %

\mypar{Self-supervised features for deepfake detection} have shown great promise.
In the \textbf{visual} domain, a frozen CLIP backbone with only a linear layer offers strong generalization \cite{ojha2023towards}.
As such, CLIP is arguably the most popular SSL encoder for image deepfake detection \cite{cozzolino2024cvprw,srivatsan23iccv,zhu23gendet,khan24icmr,koutlis2024leveraging,liu24cvpr,reiss2023detecting,smeu2024declip}, 
but others have explored other vision-language encoders
(\eg, BLIP2 \cite{reiss2023detecting,keit24bilora}, InstructBLIP \cite{chang23antifake})
or vision-only models
(\eg, DINO \cite{nguyen2024exploring}, MoCo \cite{you24arxiv}).
Similarly, the \textbf{audio} deepfake detection community has adopted large SSL representations, with most popular ones being Wav2Vec \cite{donas22icassp,wang22odyssey,wang22damm,tak2022odyssey,xie23interspeech,Pianese_2024,pascu24interspeech}, followed by HuBERT \cite{kheir25naacl} and WavLM \cite{guo2024icassp,combei2024wavlm}.
For \textbf{audio-visual} deepfake detection, AV-HuBERT representations have shown strong performance not only in a fully supervised paradigm \cite{shahzad2023avlipsyncleveragingavhubertexploit},
but also in zero-shot \cite{reiss2023detecting,liang24speechforensics} or unsupervised \cite{Smeu_2025_CVPR} settings.

\mypar{Audio-visual deepfake detection} typically exploits cross-modal inconsistencies.
Early methods targeted semantic mismatches, such as phoneme–viseme inconsistencies \cite{agarwal2020detecting} or emotional incongruences between speech and facial expressions \cite{mittal2020emotions}.
More recent work detects forgeries by measuring misalignment between dense audio-visual representations \cite{feng23cvpr,Smeu_2025_CVPR,liang24speechforensics,li2024zero}.
These representations are typically learned in a self-supervised manner, either trained from scratch \cite{oorloff2024avff,zhao2022self,wu2025hola,feng23cvpr} or obtained from pretrained models \cite{koutlis2024dimodiff,knafo2022fakeout,liang24speechforensics,reiss2023detecting}.
In contrast to these studies, we provide a comprehensive evaluation across multiple representations rather than focusing on a single one.

\section{Methodology}
\label{sec:methodology}

Deepfake detection is a binary classification task that attempts to map an input video $\xb$ to a binary label $y$
(1 if the video is fake and 0 otherwise).
We build deepfake detection models that rely mostly on self-supervised features and learn a minimal amount of parameters on top of these features. %
This approach is related to linear probing \cite{alain2017,hupkes2018,belinkov2022a}
and it allows us to assess the quality of the representations in a comparable setting.
While more complex models could be explored for downstream performance, prior work shows that simple linear models often suffice \cite{ojha2023towards,pascu24interspeech}.

The model has three steps:
1. extract locally temporal features (e.g., an embedding $\phi(\xb)_t$ for each frame $t$);
2. apply a learnable linear classifier $\wb$;
3. aggregate the predictions using a pooling function (such as log-sum-exp).
Formally, the per-video score $s$  is defined as:
\begin{equation}
    s(\xb; \wb) = \log \sum_t \exp \big\{ \wb^{\intercal} \phi(\xb)_t \big\}
\end{equation}
Since the log-sum-exp approximates the max function, the model learns to predict that a video is fake if only a single region of the model is fake.
Note that even if the model uses locally temporal features, many of the representations we test encode global temporal information (see \cref{tab:models}).
The parameters of the linear layer are trained by minimizing the cross-entropy loss on video-level labels.
See App.~\ref{sec:app-implementation-details} for more implementation details.

\subsection{Explanations}
\label{subsec:explanations}

To understand how deepfake detection classifiers make decisions,
we generate both temporal and spatial explanations.

\mypar{Temporal explanations.}
Since the pooling function (log-sum-exp) is a simple transformation of its inputs, the final video-level prediction can be regarded as an aggregation of local frame-level predictions.
We therefore compute per-frame scores $s_t = \wb^\intercal \phi(\xb)_t$ to measure which time segments contribute most to the final prediction.

\mypar{Spatial explanations.}
Given that the per-frame classifier is linear, we can further decompose each frame-level decision into patch-level contributions.
If the per-frame feature is computed by averaging patch features, we can propagate the linear classifier down to the patch level \cite{zhou2016cvpr}.
If non-linear aggregation is used, we instead apply Grad-CAM \cite{selvaraju2017grad} to obtain patch-level relevance maps.

\mypar{Evaluation.}
We compare the resulting temporal and spatial explanations against the annotated extent of local manipulations or human annotations.
Since the classifiers are trained only with video-level supervision, this comparison also serves as a form of weakly-supervised localization.

\subsection{Proxy tasks: Anomaly detection}
\label{subsec:proxy-tasks}

Instead of training a binary classifier, which risks latching onto spurious features, we explore two anomaly detection tasks that rely on real data only: next-token prediction and audio-video synchronization.
These are proxy tasks because they do not address deepfake detection directly;
rather, they model the distribution of real data, and assume that deviations from this distribution indicate fakes.
Both approaches have shown promising performance %
\cite{feng23cvpr,Smeu_2025_CVPR}.

\mypar{Next-token prediction}
models the probability of the next frame's representation $\xb_t$ given the previous frames $\xb_1,\dots,\xb_{t-1}$.
The assumption is that frames that cannot be predicted well are more likely to indicate manipulations.
We use a decoder-only Transformer trained with the mean squared error on real videos.
At test time, the model predicts each frame given its corresponding history,
and we obtain a per-video fakeness score as the maximum frame-level mean squared error.
The models' architecture has 4 layers each containing 4 heads, a feature dimension of 512 and a feedforward dimension of 1024.
In order to match the input encoding dimension, projection layers are applied before and after the Transformer.

\mypar{Audio-video synchronization}
models how well the audio and video frame-level representations match.
The assumption is that mismatches between the two modalities indicate manipulations.
We use an alignment network $\Phi$ \cite{Smeu_2025_CVPR}, where L2-normalized
audio features $\rva$ and visual features $\rvv$ are concatenated and passed through a four-layer MLP with Layer Normalization and ReLU activations.
The network is trained to maximize the probability that an audio frame $\rva_i$ aligns with its corresponding video frame $\rvv_i$, rather than with neighboring frames $ N(i)$:
\begin{equation}
p(\rvv_i \mid \rva_i) = \frac{\exp\left(\Phi(\rva_i, \rvv_i)\right)}{\sum_{k \in N(i)} \exp\left(\Phi(\rva_i, \rvv_k)\right)}.
\end{equation}
At test time, the per-frame alignment scores $\Phi(\rva_i, \rvv_i)$ are inverted to estimate fakeness, and then pooled across the video using a log-sum-exp operator to produce the final detection score.

\section{Models}
\label{sec:models}

\begin{table*}
    \centering
    \small
    \begin{tabularx}{\linewidth}{X ll lll rr}
    \toprule
    & \multicolumn{2}{c}{Input}
    & \multicolumn{3}{c}{Pretraining} & & \\
    \cmidrule(lr){2-3}
    \cmidrule(lr){4-6}
    Model  
    & \multicolumn{1}{c}{Modality} 
    & \multicolumn{1}{c}{Context} 
    & \multicolumn{1}{c}{Modality} 
    & \multicolumn{1}{c}{Content} 
    & \multicolumn{1}{c}{Dataset}  
    & Params. 
    & Dim. \\
    \midrule
    \multicolumn{8}{l}{\textit{Audio features}} \\
    Wav2Vec XLS-R 2B & audio & full & audio          & speech        & MLS and others            & 2159M & 1920 \\
    Auto-AVSR (ASR)  & audio & full & audio          & speech        & LRS3                      &  243M & 768 \\
    AV-HuBERT (A)    & audio & full & multimodal     & lips + speech & VoxCeleb, LRS3            &  310M & 1024 \\
    BRAVEn (A)       & audio & full & multimodal     & lips + speech & VoxCeleb2 (en), LRS3, AVS &  332M & 1024 \\
    \midrule
    \multicolumn{8}{l}{\textit{Visual features}} \\
    CLIP VIT-L/14    & visual & frame & visual (images) & generic        & WebImageText    & 303M & 768 \\
    FSFM             & visual & frame & visual (video)  & faces          & VGGFace2        &  86M & 768 \\
    Video-MAE-large  & visual & chunk & visual (video)  & generic        & Kinetics-400    & 303M & 1024 \\
    Auto-AVSR (VSR)  & visual & full  & visual (video)  & lips           & LRS3            & 250M & 768 \\
    AV-HuBERT (V)    & visual & full  & multimodal      & lips + speech   & VoxCeleb, LRS3 & 322M & 1024 \\
    BRAVEn (V)       & visual & full  & multimodal      & lips + speech   & VoxCeleb2 (en), LRS3, AVS & 339M & 1024 \\
    \midrule
    \multicolumn{8}{l}{\textit{Audio-visual features}} \\
    Auto-AVSR        & multimodal & full & multimodal      & lips + speech & LRS3            & 443M & 768 \\
    AV-HuBERT        & multimodal & full & multimodal      & lips + speech & VoxCeleb, LRS3 & 322M & 1024 \\
    \bottomrule
    \end{tabularx}
    \caption{%
    Overview of the models studied in the paper.
    The modality from which features are extracted (``input modality'') can differ from the modality used at training (``pretraining modality'').
    ``Context'' indicates whether the embeddings are contextual (extracted from the full input---audio or video) or local (extracted from a frame or chunk).
    }
    \label{tab:models}
\end{table*}

We study a wide range of self-supervised representation: from models trained on audio-visual data to models trained on visual-only or audio-only data.
\cref{tab:models} summarizes the models used in the paper.
The exact model checkpoints are in App.~\ref{sec:model-checkpoints}.

\subsection{Vision-only encoders}

We consider three models whose pretraining data range from faces to generic images or video.
\textbf{CLIP} \cite{radford2021learning} is an image-text model trained on general images collected from the internet.
We use the ViT/L-14 model, and take the CLS token produced by the image encoder.
\textbf{FSFM} \cite{wang2024fsfm} is a foundation model trained on faces for reconstruction.
FSFM uses a ViT/B-14 model and produces frame-level embeddings by average pooling patch embeddings.
\textbf{VideoMAE} \cite{tong2022videomae} is a vision transformer model trained to reconstruct patches of generic video.
It extracts features from spatio-temporal patches of size 2 × 16 × 16 in a window of 16 frames and with stride 16.
We discard the last window if its length is shorter than 16 frames.

\subsection{Audio-only encoders}

\textbf{Wav2Vec XLS-R 2B} \cite{baevski2020wav2vec} is a speech model trained on 436k hours of multilingual audio.
It uses a convolutional encoder followed by a transformer, and it is trained to match contextual representations with corresponding local quantized representations.
We use the final-layer features;
these are extracted every 20 ms (50 Hz), but to match 25 FPS video, we concatenate pairs of consecutive feature vectors.

\subsection{Audio-visual encoders}

We consider three audio-visual models;
all are trained on face video and extract visual features from lip regions.
\textbf{AV-HuBERT}~\cite{shi2022learning} is a Transformer that jointly encodes audio and visual inputs and is trained with masked multimodal cluster prediction.
We obtain modality-specific representations by masking one stream:
for AV-HuBERT (A) we mask visual input, and
for AV-HuBERT (V) we mask audio input.
\textbf{BRAVEn}~\cite{haliassos2024braven} uses a two-tower Transformer architecture pretrained in a cross-modal teacher–student setup:
the student predicts from masked inputs to match the teacher's prediction from the original data.
We use the two student encoders, fine-tuned for ASR and VSR, respectively.
\textbf{Auto-AVSR}~\cite{ma2021end,ma2023auto} is an audio-visual speech recognition model.\footnote{Auto-AVSR is not an SSL model, but we include it for completeness.}
Its encoders use the Conformer architecture \cite{gulati2020conformer}.
We obtain modality-specific representations from independently trained models:
the visual-only encoder trained for VSR,
and the audio-only encoder trained for ASR.

\section{Experimental setup}
\label{sec:experimental-setup}

\subsection{Datasets}
\label{sec:datasets}

We consider four audio-visual datasets varying in terms of generation methods, manipulation scope (full or local), source domain (academic or real-world).

\mypar{FakeAVCeleb} \cite{khalid2021fakeavceleb} (FAVC) contains 500 real videos from VoxCeleb2 \cite{Chung18b} and 19.5k fake videos,
generated using face swapping methods (Faceswap \cite{korshunova2017fastfaceswapusingconvolutional}; FSGAN \cite{nirkin2019fsgansubjectagnosticface}), lip-syncing (Wav2Lip \cite{Prajwal_2020}), and voice cloning (SV2TTS \cite{jia2019transferlearningspeakerverification}).
Since the dataset does not come with a predefined split, in line with previous works we split the dataset in 70\% (training and validation) and 30\% testing samples.

\mypar{AV-Deepfake1M} \cite{cai2024avdeepfake1mlargescalellmdrivenaudiovisual} (AV1M)
comprises over one million videos of approximately 2k subjects.
The videos are locally manipulated by modifying their transcripts with ChatGPT.
Corresponding fake video segments are generated with TalkLip \cite{wang2023seeingsaidtalkingface},
while fake audio segments are generated with VITS \cite{kim2021conditionalvariationalautoencoderadversarial} or YourTTS \cite{casanova22icml}.
We subsample the original training set for training and validation data,
and evaluate on 5133 samples taken from the original validation set
(see App.~\ref{sec:app-implementation-details}).

\mypar{DeepfakeEval 2024} \cite{chandra2025deepfake} (DFE-2024) is a real-world dataset collected from 88 websites through social media and the \href{https://www.truemedia.org}{\tt TrueMedia.org} platform.
The dataset contains deepfakes circulating online in 2024 across multiple media (audio, video, images) and in 52 different languages;
the types of manipulations are unknown.
We use the video subset, totaling 45 hours, which we preprocess into single-speaker segments (see App.~\ref{sec:app-dataset}).
This process yields for 70 real and 507 fake samples for test, averaging 14.06 seconds.

\mypar{AVLips} \cite{liu2024lips} (AVL) 
contains 3.4k real and 4.2k fake samples.
The real videos are sourced from LRS3 \cite{afouras2018lrs3tedlargescaledatasetvisual}, FF++ \cite{rossler2019faceforensics} and DFDC \cite{dolhansky2020deepfake},
while fake videos are generated using MakeItTalk \cite{10.1145/3414685.3417774}, Wav2Lip \cite{Prajwal_2020}, TalkLip \cite{wang2023seeingsaidtalkingface} or SadTalker \cite{zhang2023sadtalker}.
We evaluate on the whole dataset.

\subsection{Evaluation and baselines}

\textbf{Classification.}
To measure how well fake samples can be distinguished from real ones,
we report the area under receiver operator characteristic curve (AUC).
This metric has the advantages of being threshold-independent and having a clear baseline:
a random model achieves 50\% AUC, regardless of the class distribution.
We also report average precision results in App.~\ref{app:average-precision},~\cref{tab:main-results_AP} \&~\cref{tab:ntp-sync-sup-results_AP}.
For a fair evaluation, only fake samples with both audio and visual manipulations are considered, where this information is available.

\mypar{Temporal localization.}
We evaluate how well temporal explanations align with ground-truth annotations.
For this, we consider the videos from the AV1M test set that contain at least one fake segment,
and compute a localization score per video.
The localization score is computed in terms of AUC by treating each frame as an independent sample:
the model's frame-level score serves as the prediction,
while annotated fake segment specifies the groundtruth label.
The final localization score is the average of the AUC values across all fake test videos.

\mypar{Baselines.}
To understand how much information is implicitly encoded in the architecture,
we report results of an untrained (randomly initialized) AV-HuBERT model.
To further contextualize the results, we include five state-of-the-art approaches:
AVAD \cite{feng23cvpr},
AVFF \cite{oorloff2024avff},
SpeechForensics \cite{liang24speechforensics},
AuViRe \cite{koutlis2025auvire},
RealForensics \cite{haliassos22realforensics}.
For AVFF we use the unofficial open source implementation%
\footnote{\url{https://github.com/JoeLeelyf/OpenAVFF}}
and finetune the Kinetics400 checkpoint on the same datasets as the linear probes.
For the rest we use the pretrained models.
For more details, see App.~\ref{app:baselines}.

\input{sec/6-experimental-results-v2}

\section{Conclusions}

We examined a wide array of self-supervised representations for audio-visual deepfake detection.
Many of these features, although complementary, are able to perform strongly in-domain, even on challenging locally manipulated data.
The representations that generalize best are audio-informed features;
in particular, the visual encoder of BRAVEn achieves state-of-the-art results, outperforming more complex approaches.
Audio-only features also perform strongly, but only when datasets contain speech-level manipulations.

A surprising finding was that randomly initialized features can achieve strong performance and even generalize across datasets.
This observation suggests that standard supervised evaluation can be driven by dataset shortcuts.
Indeed, by considering a multifaceted evaluation protocol (anomaly detection, temporal and spatial alignment), we are able to expose such failure modes and show that %
random features are not aligned with groundtruth manipulations.

\textbf{Limitations.}
There remains a gap towards real-word deepfake detection:
neither the tested representations nor existing approaches generalize well to such data.
Our results indicated that this limitation persists despite the representations being well aligned to semantic artifacts (temporal and spatial).
The likely reason is the difficulty and diversity of the real-world data
(\eg, missing modalities, shifts in the video domain).
Addressing this challenge will require more specialized solutions that explicitly account for this diversity.

\section*{Acknowledgments}
We thank the reviewers and area chair for their suggestions.
This work was supported in part by the EU Horizon project ELIAS (No.101120237) and by CNCS-UEFISCDI (PN-IV-P7-7.1-PTE-2024-0600 and PN-IV-P2-2.1-TE-2023-1632).

{
    \small
    \bibliographystyle{ieeenat_fullname}
    \bibliography{main}
}

\input{sec/8-supplementary-material}

\end{document}

%% file: sec/6-experimental-results-v2.tex
\input{tab-main-v3}

\section{Experimental results}
\label{sec:experimental-results}

We first evaluate how useful are the representations for both the direct task of deepfake detection,
as well as the proxy task of anomaly detection (\cref{subsec:exp-rq1}).
Then we evaluate the models in terms of what they look at when making their decision (\cref{subsec:exp-rq2}).
Finally, we quantify how different the deepfake information encoded by the representations is (\cref{subsec:exp-rq3}).

\subsection{How useful are the representations?}
\label{subsec:exp-rq1}

\cref{tab:main-results-v2} shows the main results:
linear probes results
across the selected self-supervised features and
across multiple combinations of in-domain and out-of-domain datasets.

\mypar{Many features can encode deepfake information.}
We observe strong in-domain results on the scientific datasets (cols. \mylabel{A} and \mylabel{C}),
which show that many representations are able to encode the subtle information needed to differentiate between real and fake samples.
This is especially remarkable for the AV1M dataset, which has only short temporal manipulations.
Interestingly, the deepfake information can be picked from different angles:
from audio (Wav2Vec2, AV-HuBERT (A)), from motion information (Video-MAE), from static vision content (CLIP).

\mypar{Audio-informed features generalize best.}
On FAVC and AV1M (cols. \mylabel{A:D}), audio representations perform best.
On AVLips and DFE-2024 (cols. \mylabel{E:F}), which lack or have incomplete audio manipulations, audio-informed visual representations (AV-HuBERT (V) and BRAVEn (V)) are best.
These models, albeit simple, achieve state-of-the-art performance (\cref{tab:sota}).
Among existing methods, only SpeechForensics approaches this performance;
but as discussed in App.~\ref{sec:speechforensics},
this is because it relies on similar features (AV-HuBERT) and formulates detection as anomaly detection.
We also experimented with a more powerful backend classifier (a transformer) and observed similar results (App. \ref{app:classification-head-analysis}), suggesting that features are more important than the classifier.

\input{tab-sota}

\mypar{In-the-wild data is not only different, but also harder.}
Generalization to DFE-2024 (cols. \mylabel{H:I}) is poorer than to other datasets (cols. \mylabel{B, D:F}):
the best out-of-domain result on DFE-2024 reaches only 76.0\% AUC, achieved by BRAVEn (V) (cell \mylabel{I12}).
While this gap may reflect differences between datasets (in terms of content or artifacts),
it also indicates that DFE-2024 is intrinsically more difficult \cite{muller2024harder}:
even the best in-domain result is at only 75.5\% AUC (cell \mylabel{G12}).

\mypar{Random features are not random.}
Surprisingly, results of the randomly-initialized models (rows \mylabel{1:2}) are well above random chance (50\% AUC).
This indicates that the architectures can implicitly encode information that is discriminative.
But how useful is this information really?
Results from \cref{subsec:exp-rq2} suggest that random features encode spurious cues, such as the leading silence \cite{Smeu_2025_CVPR,klemt2025deepfake},
which explains the better performance of the random audio model compared to its visual counterpart.

\input{tab-proxy-tasks-thin}

\mypar{Anomaly detection offers robustness, but only certain feature combinations work.}
\cref{tab:ntp-sync-sup-results-v2} presents deepfake detection performance for models trained on the two anomaly detection tasks (next-token prediction and synchronization), along with cross-domain supervised results.
As just seen, supervised models can perform well even when trained on features from random models.
When these random features are used for the two proxy anomaly detection tasks, performance drops sharply:
down to chance level for synchronization and to moderate values for next-token prediction.
This confirms that the unsupervised methods are more robust:
unlike supervised classifiers, they do not exploit spurious correlations that random features may inadvertently encode.
However, to achieve reasonable anomaly detection results, specific feature combinations are required.
For next-token prediction, AV-HuBERT (A) features are essential:
they perform best individually and substantially improve performance when combined with others.
For synchronization, however, AV-HuBERT (V) features are crucial;
this contrasts with CLIP or random features, which are insufficient.
The only setting that comes close to the supervised model is the synchronization model on AV-HuBERT (A) + AV-HuBERT (V).
Taken together, the results indicate that robustness in anomaly detection depends on selecting features that capture the temporal dynamics necessary to align and compare audio–visual streams effectively.

\subsection{What do models look at?}
\label{subsec:exp-rq2}

\begin{figure}
    \centering
    \includegraphics[width=\linewidth]{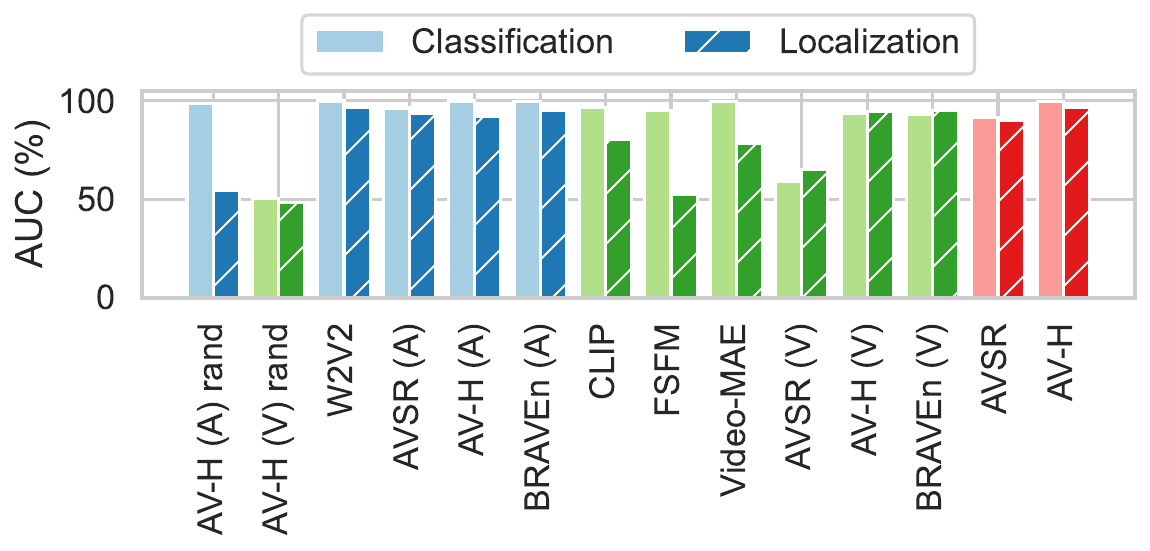}
    \caption{
        Temporal localization of explanations (solid bars) and deepfake classification (hatched bars) on AV1M.
        Colors indicate modality: blue (audio), green (visual), red (audio-visual).
    }
    \label{fig:det-vs-loc}
\end{figure}

\mypar{Localization of temporal explanations: Audio models pick leading silence, but also useful artifacts.}
\cref{fig:det-vs-loc} shows the alignment between the temporal explanations and local manipulations from AV1M.
We see that many features still produce strong results (localization performance is close to that of classification),
indicating that linear probes extract reasonable information from the representations.
Large performance drops appear only for the random models, confirming that they rely on spurious cues, and for FSFM, where inspection did not reveal a visible bias.
Audio models show strong performance,
but qualitative examples in \cref{fig:temporal-explanations} indicate that they do attend to leading silence.
Despite this, they also focus on the manipulated regions, retaining good localization performance;
interestingly, Wav2Vec2 seems to select transition boundaries. 
Among the four models shown, AV-HuBERT (V) produces the cleanest predictions, while CLIP is the noisiest---%
its activations often fire on blurry frames, although this is difficult to quantify since scores fluctuate even between adjacent, visually similar frames.

\input{fig-temporal-explanations}

\input{fig-spatial-explanations}

\mypar{Spatial alignment to human annotations: CLIP attends to facial artifacts, similarly to humans.}
We saw that the models find temporarily modified regions.
Do they also look at the same artifacts as humans?
To answer this question, we analyze the recently introduced ExDDV dataset \cite{hondru2025exddv},
which provides click annotations indicating where humans identified generation artifacts.
We compare these annotations with explanations produced by a CLIP-based model trained on the ExDDV training set
(this model achieves 71.3\% AUC on the ExDDV test set).
We generate explanations for the fake test videos using GradCAM applied to the final LayerNorm layer of the CLIP visual encoder.
We quantify the human-machine alignment as the mean absolute error (MAE) between the relative coordinates of human click annotations and the maximum values in the GradCAM attention maps.
To contextualize these results, we compare against several baselines:
random position within each frame, frame center, face center, and a predictive click model trained on click annotations (the ViT model from \cite{hondru2025exddv}).

Fig.~\ref{fig:spatial-explanations} (left) presents quantitative results:
MAE as a function of the minimum fakeness score.
We observe that error decreases with the minimum fakeness score, suggesting better alignment for confident predictions (these predictions are also correct, since we consider only fake samples).
The variance increases correspondingly due to fewer samples exceeding the higher score thresholds.
Compared to baseline methods, the explanations achieve better alignment than frame center (0.117 MAE) or random locations (0.270 MAE, not shown).
However, alignment remains lower than what a predictive model achieves (0.055 MAE). Notably, the predictive click model performs only marginally better than face center prediction (0.058 MAE), suggesting that human annotations may not contain substantially more localization information beyond indicating that artifacts occur somewhere on the face.
Fig.~\ref{fig:spatial-explanations} (right) shows qualitative results.
In these samples, most model predictions concentrate on the forehead region, while human annotations focus on eyes and lips.
Crucially, the model appears to avoid relying on spurious background features, with explanations consistently concentrating on facial regions.

\subsection{How complementary are the representations?}
\label{subsec:exp-rq3}

\begin{figure} %
    \centering
    \includegraphics[width=\linewidth]{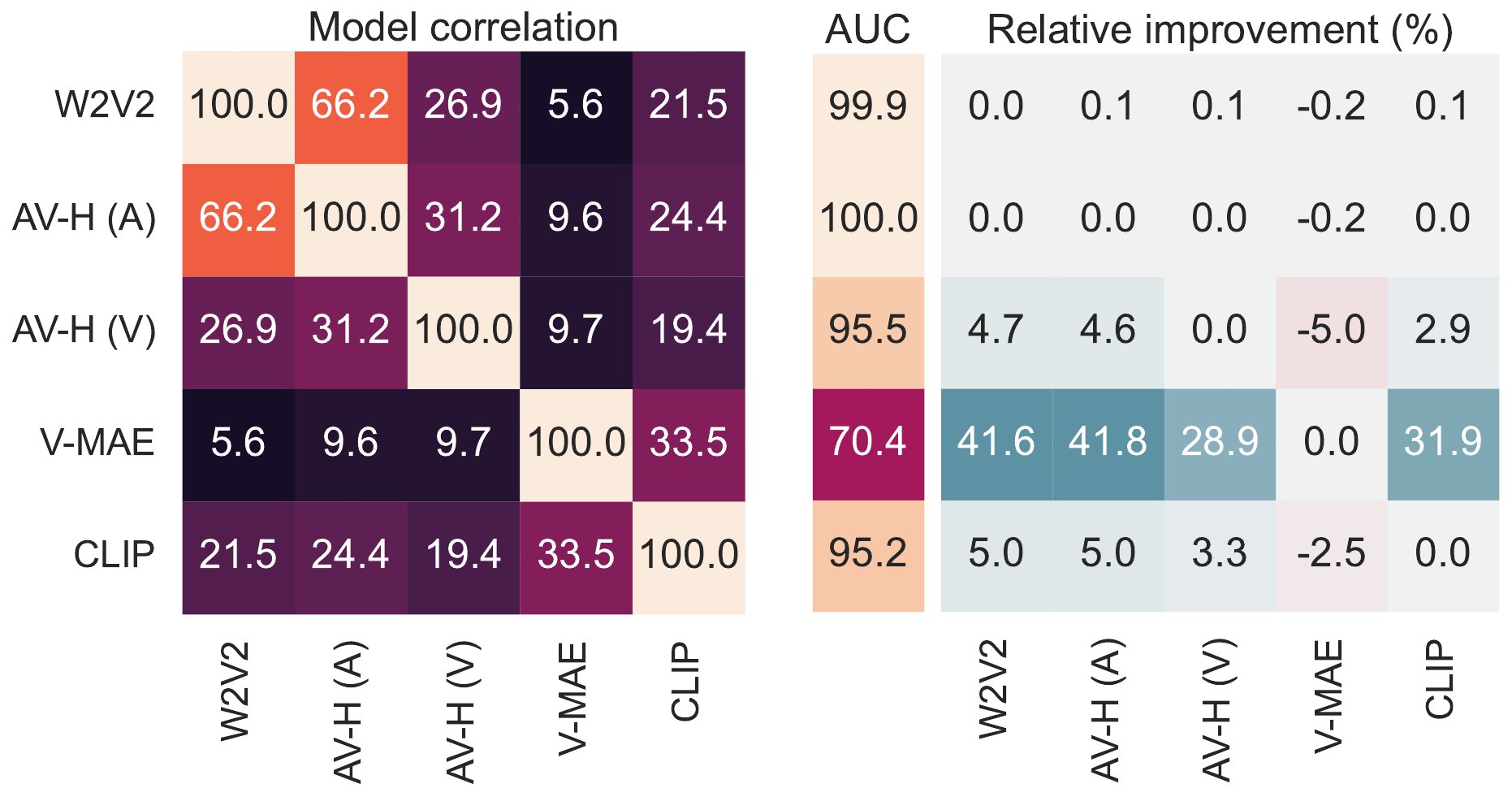}
    \caption{
        Correlations between models (left) and downstream performance (right).
        The downstream performance is presented in absolute values for the unimodal models (AUC column) and
        as relative improvement for feature combinations.
        Training was done on AV1M, testing on FAVC.
    }
    \label{fig:model-combination}
\end{figure}

To understand how the different self-supervised representations relate, we examine two aspects.
First, we measure the correlation between predictions produced by pairs of models.
This evaluation provides insight into the similarity of their learned decision boundaries.
Second, we evaluate downstream performance when combining multiple models.
This offers a more direct assessment of the feature combination effectiveness toward our final goal.
Results on FAVC are in Fig.~\ref{fig:model-combination} and on DFE-2024 in App.~\ref{app:combinations-of-features}, Fig.~\ref{fig:model-comb-av1m-dfeval}.

\mypar{Vision models are more complementary than audio ones.}
For each pair of models, we generate predictions on a shared test dataset and calculate the Pearson correlation coefficient between their outputs.
\cref{fig:model-combination} (left) shows that the cross-model correlations are generally weak to moderate;
this suggests that the embeddings encode different information.
The strongest correlations occur within modalities:
in particular, audio models (AV-HuBERT (A) and Wav2Vec2) show highest correlation with each other,
stronger than the vision counterpart (CLIP and VideoMAE).
Notably, AV-HuBERT (V) correlates more strongly with audio models than with other video models.
This happens because AV-HuBERT (V) focuses solely on lip movements and is trained jointly with the audio component.

\mypar{Complementarity helps downstream performance, but there are exceptions.}
The largest gains from feature combination occur for Video-MAE.
This is expected, as Video-MAE has the lowest performance and thus has the most room for improvement.
Also, unsurprisingly, Video-MAE benefits most from audio features, which are both stronger and more complementary than the visual ones.
However, among similarly performing features, Video-MAE benefits more from CLIP (with which is more aligned) than from AV-HuBERT (V) (with which is more complementary).
This suggests that the impact of feature combination is more nuanced than solely their complementarity.

%% file: tab-main-v3.tex
\begin{table*}
    \definecolor{mycolor}{HTML}{d4a373}
    \newcolumntype{i}{r}
    \newcolumntype{o}{r}
    \centering
    \small
    \tabcolsep 4.5pt
    \begin{tabularx}{\linewidth}{r X c io io oo ioo o}
        \toprule
        &
        &
        & \multicolumn{1}{c}{\ii{A}}
        & \multicolumn{1}{c}{\ii{B}}
        & \multicolumn{1}{c}{\ii{C}}
        & \multicolumn{1}{c}{\ii{D}}
        & \multicolumn{1}{c}{\ii{E}}
        & \multicolumn{1}{c}{\ii{F}}
        & \multicolumn{1}{c}{\ii{G}}
        & \multicolumn{1}{c}{\ii{H}}
        & \multicolumn{1}{c}{\ii{I}}
        & \multicolumn{1}{c}{\ii{J}}
        \\
        &
        &
        & \multicolumn{2}{c}{Test on FAVC}
        & \multicolumn{2}{c}{Test on AV1M}
        & \multicolumn{2}{c}{Test on AVLips}
        & \multicolumn{3}{c}{Test on DFE-2024}
        \\
        \cmidrule(lr){4-5}
        \cmidrule(lr){6-7}
        \cmidrule(lr){8-9}
        \cmidrule(lr){10-12}
        &
        &
        & \multicolumn{1}{c}{FAVC} & \multicolumn{1}{c}{AV1M}
        & \multicolumn{1}{c}{AV1M} & \multicolumn{1}{c}{FAVC}
        & \multicolumn{1}{c}{FAVC} & \multicolumn{1}{c}{AV1M}
        & \multicolumn{1}{c}{DFE} & \multicolumn{1}{c}{FAVC} & \multicolumn{1}{c}{AV1M}
        \\
        &
        &
        & \multicolumn{1}{c}{$\downarrow$}
        & \multicolumn{1}{c}{$\downarrow$}
        & \multicolumn{1}{c}{$\downarrow$}
        & \multicolumn{1}{c}{$\downarrow$}
        & \multicolumn{1}{c}{$\downarrow$}
        & \multicolumn{1}{c}{$\downarrow$}
        & \multicolumn{1}{c}{$\downarrow$}
        & \multicolumn{1}{c}{$\downarrow$}
        & \multicolumn{1}{c}{$\downarrow$} 
        & \multicolumn{1}{c}{mean}
        \\
        & Model            
        & Modality
        & \multicolumn{1}{c}{FAVC} & \multicolumn{1}{c}{FAVC}
        & \multicolumn{1}{c}{AV1M} & \multicolumn{1}{c}{AV1M}
        & \multicolumn{1}{c}{AVLips} & \multicolumn{1}{c}{AVLips}
        & \multicolumn{1}{c}{DFE} & \multicolumn{1}{c}{DFE} & \multicolumn{1}{c}{DFE}& \multicolumn{1}{c}{OOD}
        \\
        \midrule
        \ii{1}   & AV-HuBERT (A) random                           & audio  & \bf 99.8 & \bf 97.8 & \bf 98.8 & \bf 92.0 & \gray 50.9     & \gray 51.9     & \gray 53.4     & \gray 41.9     & \gray 46.4     & \bf 63.5 \\
        \ii{2}   & AV-HuBERT (V) random                           & visual & 83.8     & 52.5     & 50.2     & 49.3     & \bf 60.4       & \bf 53.8       & \bf 56.8       & \bf 60.6       & \bf 54.0       & 55.1     \\
        \midrule 
        \ii{3}   & Wav2Vec2                                       & audio  & \bf 100 & 99.9     & \bf 100  & 96.6     & \gray 51.3     & \gray 56.3     & \gray 58.7     & \gray 62.3     & \gray \bf 58.6 & \bf 70.8 \\
        \ii{4}   & Auto-AVSR (ASR)                                & audio  & 99.7    & 76.0     & 96.4     & 50.3     & \gray \bf 52.9 & \gray 49.6     & \gray 63.5     & \gray 49.4     & \gray 47.5     & 54.3     \\
        \ii{5}   & AV-HuBERT (A)                                  & audio  & \bf 100 & \bf 100  & \bf 100  & \bf 99.0 & \gray 50.0     & \gray \bf 57.2 & \gray 65.8     & \gray 49.1     & \gray 48.3     & 67.3     \\
        \ii{6}   & BRAVEn (A)                                     & audio  & \bf 100 & 99.7     & 99.9     & 89.4     & \gray 44.5     & \gray 47.6     & \gray \bf 67.8 & \gray \bf 64.2 & \gray 54.6     & 66.7     \\
        \midrule 
        \ii{7}   & CLIP VIT-L/14                                  & visual & 99.8    & 95.2     & 96.5     & \bf 71.1 & 60.3           & 53.3           & 73.9           & 55.6           & 43.5           & 63.2     \\
        \ii{8}   & FSFM                                           & visual & 97.1    & 40.9     & 95.3     & 52.7     & 84.3           & 36.8           & 71.7           & \bf 71.8       & 43.5           & 55.0     \\
        \ii{9}   & Video-MAE-large                                & visual & \bf 100 & 70.4     & \bf 99.8 & 60.0     & 71.3           & 47.2           & 54.5           & 45.6           & 39.3           & 55.6     \\
        \ii{10}  & Auto-AVSR (VSR)                                & visual & 97.8    & 77.5     & 59.0     & 51.3     & 83.3           & 70.1           & 64.3           & 48.7           & 56.1           & 64.5     \\
        \ii{11}  & AV-HuBERT (V)                                  & visual & \bf 100 & 95.5     & 93.7     & 64.1     & 98.3           & 90.5           & 72.1           & 63.7           & 67.7           & 80.0     \\
        \ii{12}  & BRAVEn (V)                                     & visual & \bf 100 & \bf 98.8 & 93.0     & 66.3     & \bf 98.9       & \bf 96.7       & \bf 75.5       & 70.8           & \bf 76.0       & \bf 84.6 \\
        \midrule 
        \ii{13}  & Auto-AVSR                                      & audio-visual & 94.7    & 68.3     & 91.6     & 53.2     & 59.6           & 54.6           & 61.2            & 43.0           & 49.2           & 54.7     \\
        \ii{14}  & AV-HuBERT                                      & audio-visual & \bf 100 & \bf 99.5 & \bf 99.9 & \bf 94.5 & \bf 78.5       & \bf 84.4       & \bf 70.4        & \bf 58.2       & \bf 54.3       & \bf 78.2 \\
        \bottomrule
    \end{tabularx}
    \caption{%
        Area under curve (AUC, \%) of linear probes trained on different representations and evaluated in (columns A, C, G) and out of domain (other columns).
        AVLips and DFE-2024 lack or have incomplete audio manipulations, hence the gray values for the audio features.
    }
    \label{tab:main-results-v2}
\end{table*}

%% file: tab-sota.tex
\begin{table}[t]
\footnotesize
\setlength{\tabcolsep}{3pt}
\centering
\begin{tabular}{lcrrrrrr}
\toprule
& 
& \multicolumn{4}{c}{Test dataset}
& \multicolumn{2}{c}{Average} \\
\cmidrule(lr){3-6}
\cmidrule(lr){7-8}
Method & Sup? & AV1M & FAVC & AVL & DFE & All4 & Last3 \\
\midrule
\multicolumn{8}{l}{\textit{Self-supervised features (best in each modality) and linear probing}} \\
Wav2Vec2 (A)             & \cmark & \bf 100   & \bf 99.9 & 56.3     & 58.6     & 78.7     & 71.6\\
BRAVEn (V)               & \cmark & 93.0      & 98.8     & \bf 96.7 & \bf 76.0 & \bf 91.1 & \bf 90.5 \\
AV-HuBERT (AV)           & \cmark & 99.9      & 99.5     & 84.4     & 54.3     & 84.5     & 79.4 \\
\midrule
\multicolumn{8}{l}{\textit{State-of-the-art approaches: Supervised and unsupervised}} \\
AVFF \cite{oorloff2024avff}                              & \cmark & 98.0    & 89.2    & 53.9     & 58.1     & 74.8     & 67.1 \\
AuViRe$^\dagger$ \cite{koutlis2025auvire}                & \cmark & \bf 100 & 88.1    & 72.4     & 63.8     & 81.1     & 74.7 \\
RealForensics$^\ddagger$ \cite{haliassos22realforensics} & \cmark & 60.7    & 89.1    & 76.8     & 74.4     & 75.2     & 80.1 \\
AVAD \cite{feng23cvpr}                                   &        & 52.9    & 95.2    & 73.2     & 64.8     & 71.5     & 77.7 \\
SpeechForensics \cite{liang24speechforensics}            &        & 68.1    & \bf 100 & \bf 92.7 & \bf 75.6 & \bf 84.1 & \bf 89.4 \\
\bottomrule
\end{tabular}
\caption{%
    Comparison to state of the art.
    AUC (\%) when training on the AV1M dataset (23k samples).
    We report average score across all datasets (All4) and without AV1M (Last3).
    Legend:
    $\dagger$ trained on the full AV1M training set (746k samples);
    $\ddagger$ %
    trained on FF++~\cite{rossler2019faceforensics}.
}
\label{tab:sota}
\end{table}

%% file: tab-proxy-tasks-thin.tex
\begin{table}
    \centering
    \setlength{\tabcolsep}{3pt}
    \footnotesize
    \begin{tabularx}{\linewidth}{X rrr rrr}
        \toprule
                            & \multicolumn{3}{c}{AV1M} & \multicolumn{3}{c}{FAVC} \\
                            \cmidrule(lr){2-4}
                            \cmidrule(lr){5-7}
        Model               & Sup. & NTP       & Sync.     & Sup. & NTP       & Sync. \\
        \midrule
        \multicolumn{5}{l}{\textit{Single features}} \\
        AV-HuBERT (A)       & \bf 99.0 & \bf 90.6 & \na       & \bf 100  & \bf 80.5 & \na \\
        Wav2Vec2            & 96.6     & 56.6     & \na       & 99.9     & 59.4     & \na \\
        AV-HuBERT (V)       & 64.1     & 46.1     & \na       & 95.5     & 55.3     & \na \\
        CLIP                & 71.1     & 47.3     & \na       & 95.2     & 60.2     & \na \\
        \midrule
        \multicolumn{5}{l}{\textit{Combination of features}} \\
        AV-H (A + V) random  & 74.0 & 64.4      & 50.0      & 95.0    & 75.4      &  50.3 \\
        AV-H (A + V)         & 97.2 & 84.5     & \bf 87.3 & \bf 100 & \bf 91.2 & \bf 96.3 \\
        AV-H (A) + CLIP      & \bf 99.0 & \bf 86.9 & 50.0     & \bf 100 & 79.6     & 54.4 \\
        W2V2 + AV-H (V)      & 96.2 & 60.6     & 86.5     & \bf 100 & 79.4     & 94.6 \\
        W2V2 + CLIP          & 97.1 & 57.2     & 49.7     & \bf 100 & 69.8     & 21.7 \\
        \bottomrule
    \end{tabularx}
    \caption{%
        AUC performance (\%) on AV1M and FAVC of supervised (sup.) and anomaly detection models: next-token prediction (NTP) and audio-video synchronization (sync.). 
        Supervised models are trained cross-domain (FAVC$\to$AV1M and AV1M$\to$FACV, respectively),
        while anomaly detection models are trained on real data only (a subset of VoxCeleb).
        Supervised models of feature combinations use late fusion (average of predictions).
    }
    \label{tab:ntp-sync-sup-results-v2}
\end{table}

%% file: fig-temporal-explanations.tex
\begin{figure}
    \centering
    \setlength{\tabcolsep}{1pt}
    \newcommand\htc{\hskip 3pt}
    \newcommand\myplot[1]{\includegraphics[width=4.0cm]{#1}}
    \newcommand\myimg[1]{\includegraphics[width=1.3cm, height=1.3cm]{#1}}
    \newcommand\myt[1]{\footnotesize $t$: #1}
    \newcommand\mycombp[2]{\footnotesize #1 $\cdot$ score: #2}
    \newcommand\mylabelx[1]{\scriptsize \sf #1}
    \newcommand\myproba[1]{\footnotesize $p$: #1}
    \begin{tabular}{ccc @{\htc} ccc}
        \multicolumn{3}{c@{\htc}}{\mycombp{AV-HuBERT (A)}{13.53}} & \multicolumn{3}{c@{\htc}}{\mycombp{Wav2Vec2}{16.54}} \\ 
        \multicolumn{3}{c@{\htc}}{\myplot{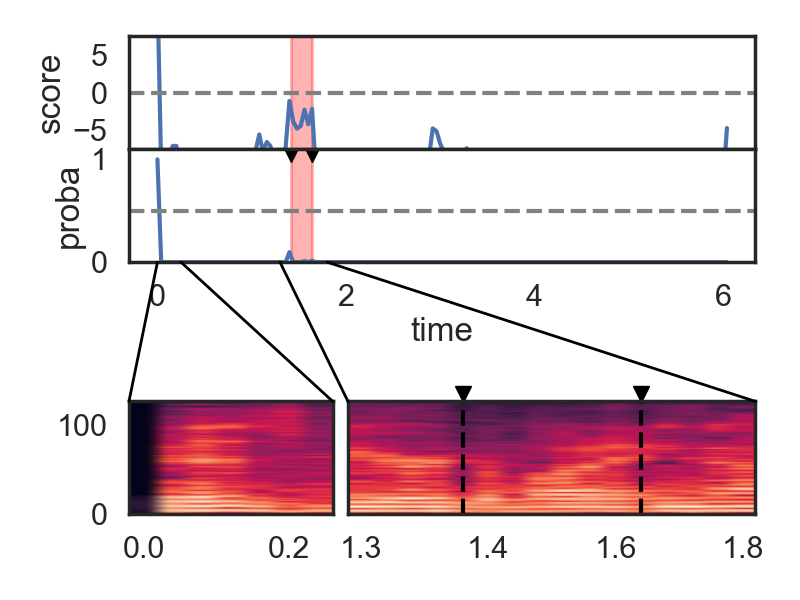}} & \multicolumn{3}{c@{\htc}}{\myplot{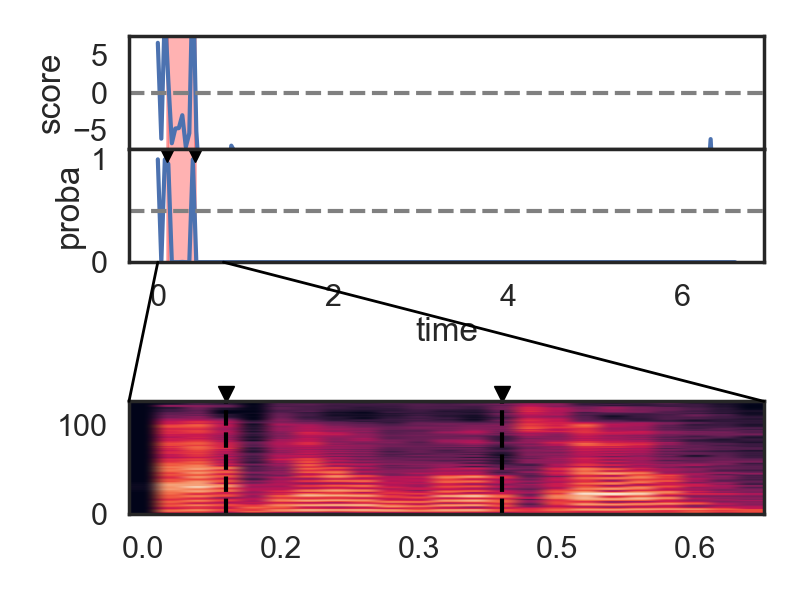}} \\ 
        \multicolumn{3}{c@{\htc}}{\mycombp{AV-HuBERT (V)}{9.54}} & \multicolumn{3}{c@{\htc}}{\mycombp{CLIP}{7.17}} \\ 
        \multicolumn{3}{c@{\htc}}{\myplot{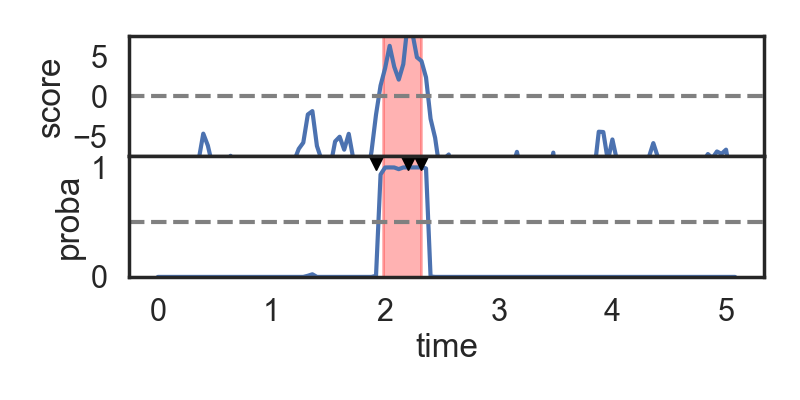}} & \multicolumn{3}{c@{\htc}}{\myplot{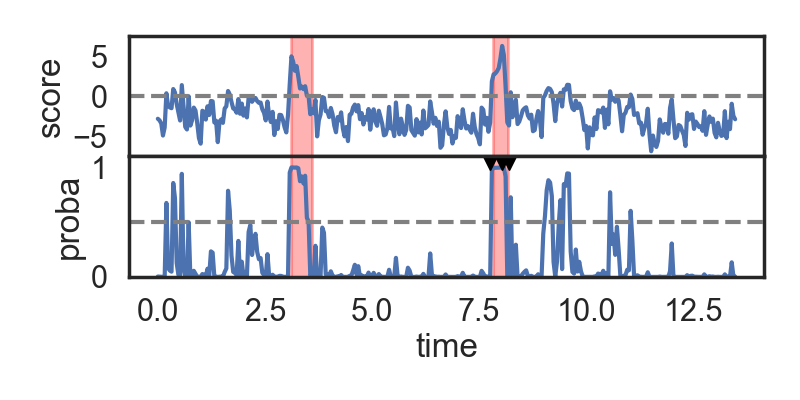}} \\ [-0.3cm]
        \myt{1.9s} & \myt{2.2s} & \myt{2.3s} & \myt{7.8s} & \myt{8.0s} & \myt{8.2s} \\ 
        \myimg{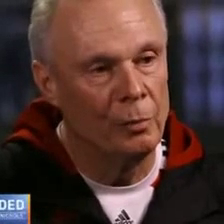} & \myimg{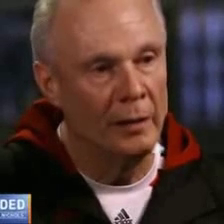} & \myimg{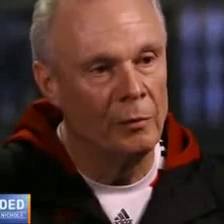} & \myimg{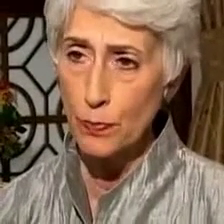} & \myimg{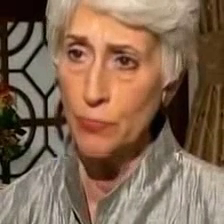} & \myimg{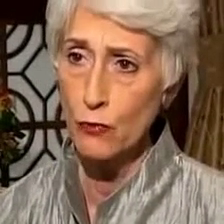} \\
        \mylabelx{real} & \mylabelx{fake} & \mylabelx{real} & \mylabelx{real} & \mylabelx{fake} & \mylabelx{real} \\ 
        \myproba{0.0} & \myproba{1.0} & \myproba{1.0} & \myproba{0.0} & \myproba{1.0} & \myproba{0.0} \\
   \end{tabular}
    \caption{%
        Temporal explanations of the top video predictions for four SSL representations. %
        The predictions are given in terms of unnormalized scores (logits) and probabilities.
        Red regions indicate fake segments, and gray dashed lines correspond to the decision boundary (0.5 probability).
        For audio models we show Mel spectrograms;
        for vision models we show three frames (corresponding to the triangle markers on the line plot).
    }
    \label{fig:temporal-explanations}
\end{figure}

%% file: fig-spatial-explanations.tex
\begin{figure*}
    \centering
    \begin{minipage}{0.28\textwidth}
        \includegraphics[width=\linewidth]{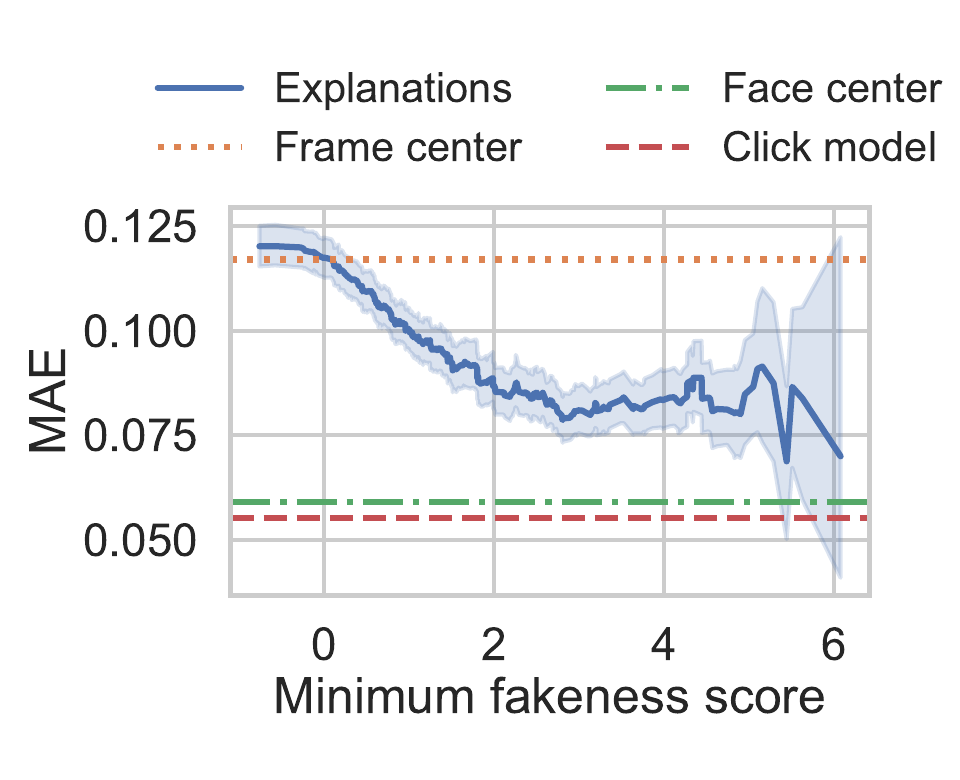}
    \end{minipage}
    \hspace{0.2cm}
    \begin{minipage}{0.61\textwidth}
        \setlength{\tabcolsep}{1pt}
        \newcommand\myimg[1]{\includegraphics[width=1.8cm, height=1.4cm]{#1}}
        \scriptsize
        \begin{tabular}{cccccc}
            score: 4.522 & score: 2.123 & score: 0.872 & score: 0.871 & score: 0.593 & score: 0.439 \\ 
            \myimg{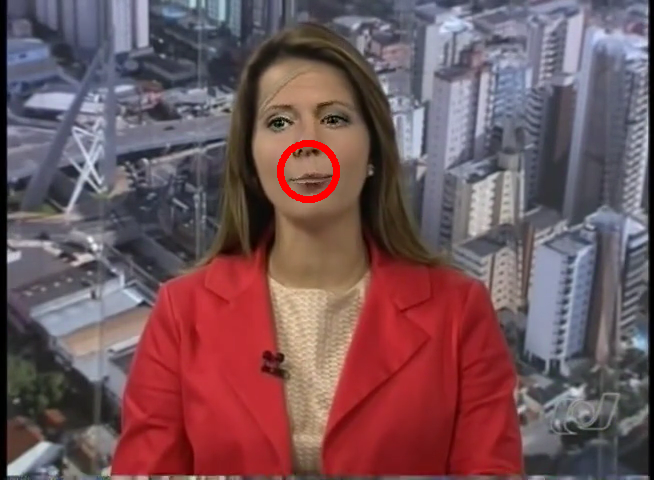} & \myimg{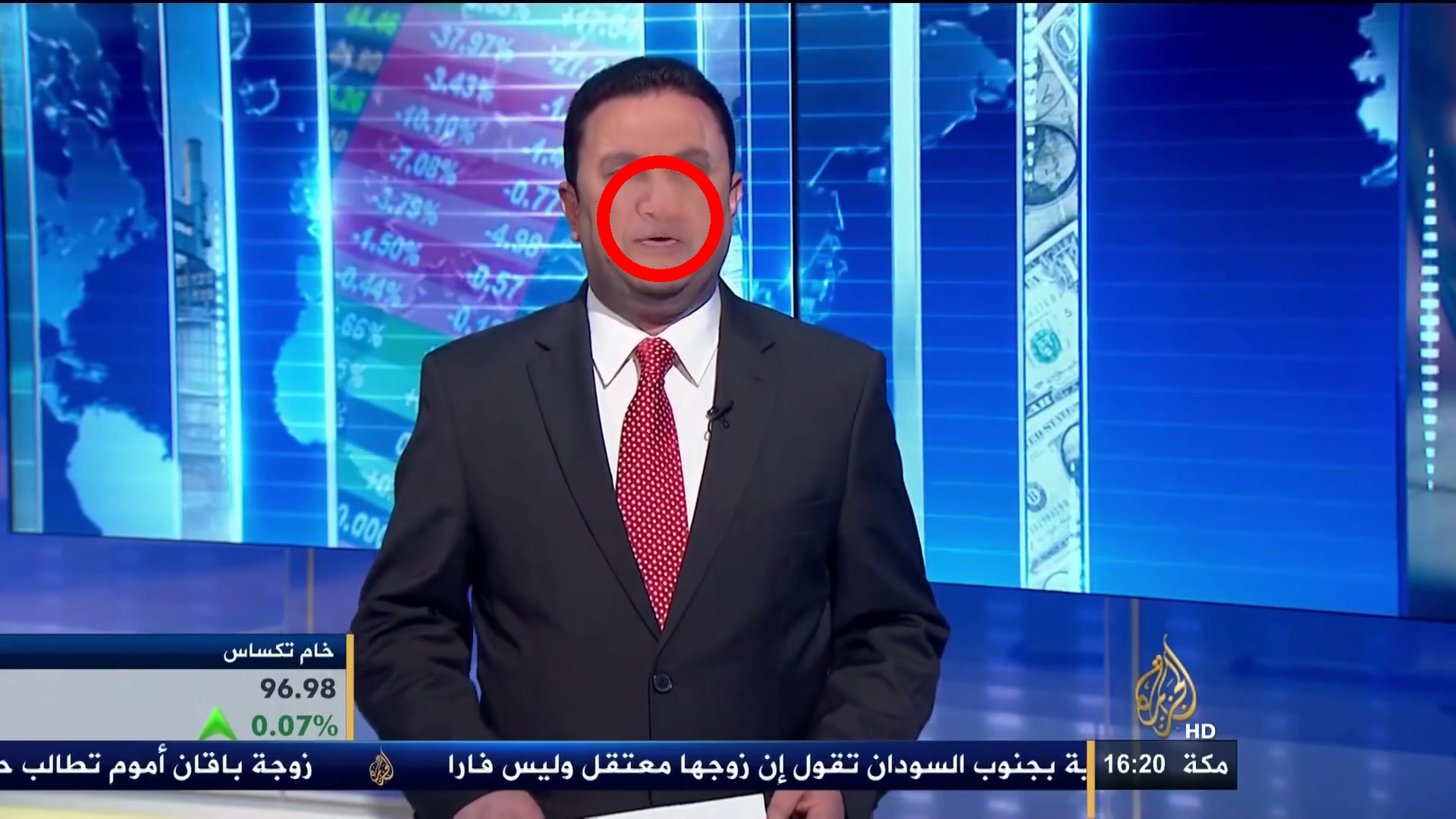} & \myimg{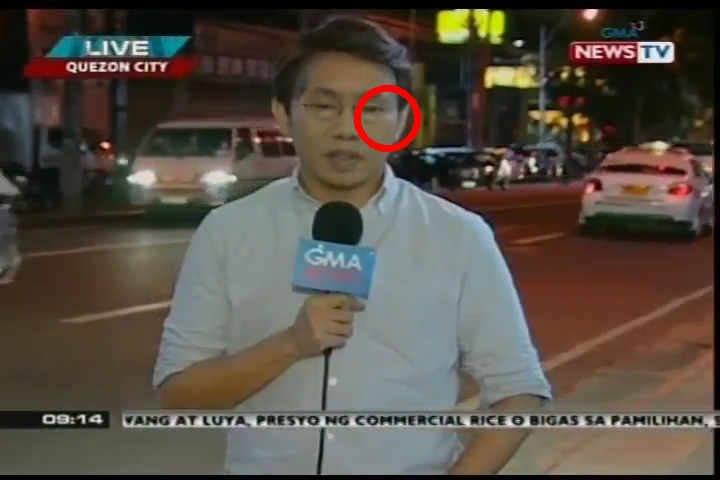} & \myimg{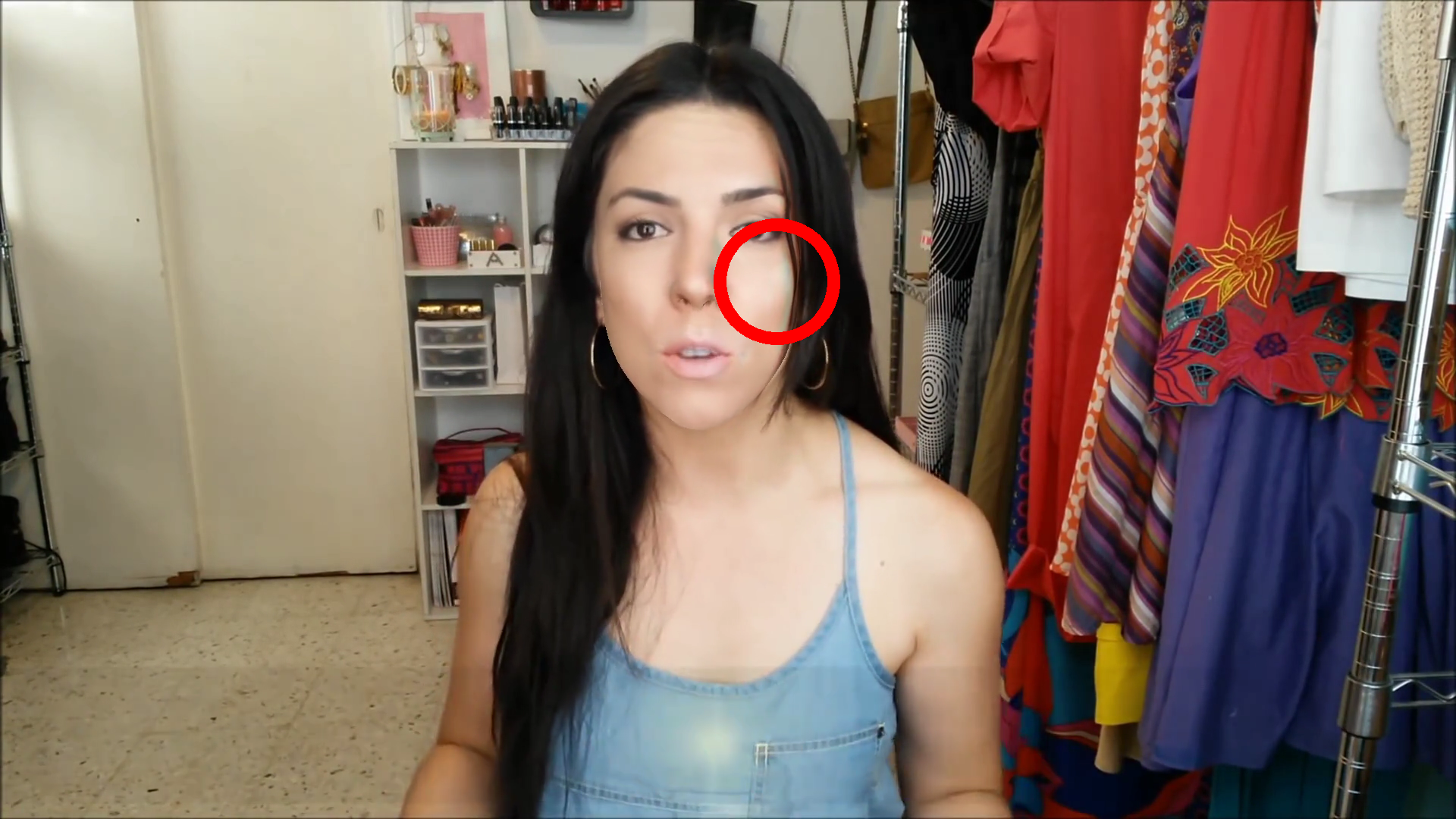} & \myimg{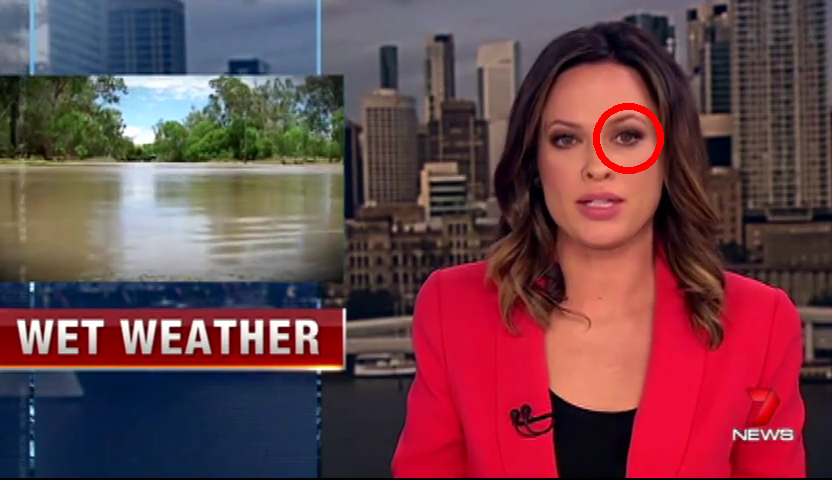} & \myimg{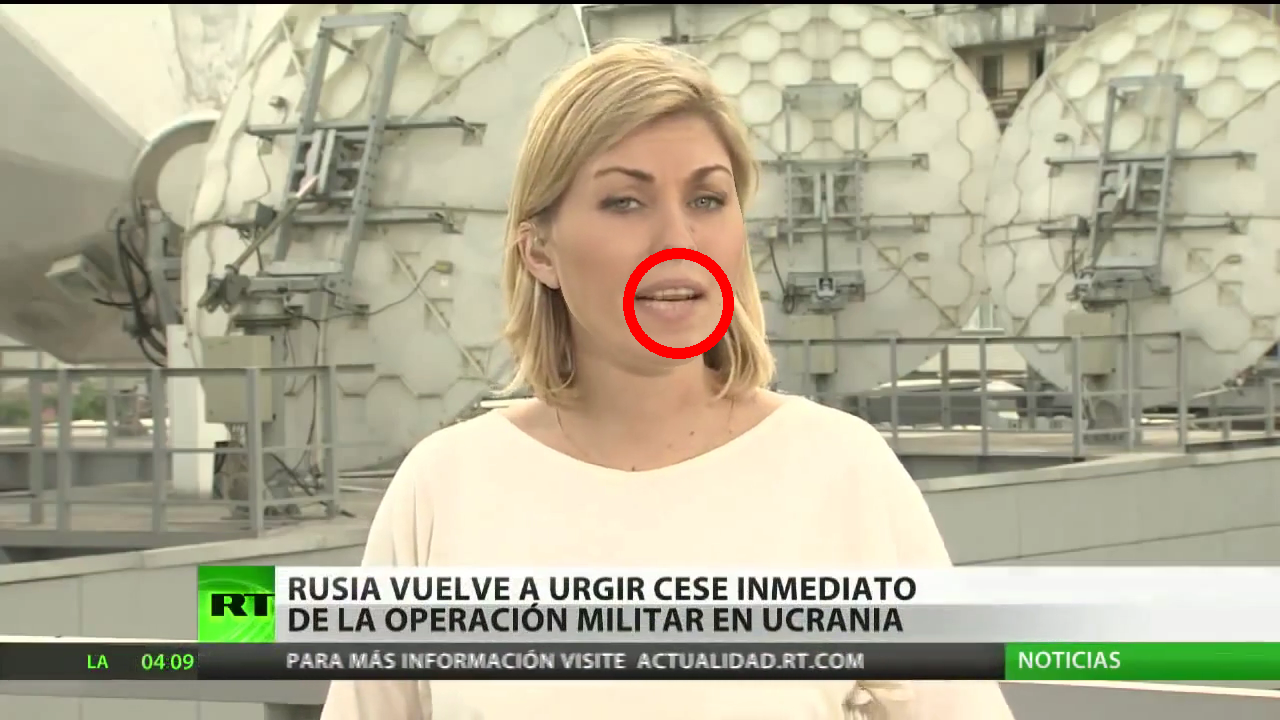} \\ 
            error: 0.120 & error: 0.039 & error: 0.126 & error: 0.009 & error: 0.240 & error: 0.108 \\ 
            \myimg{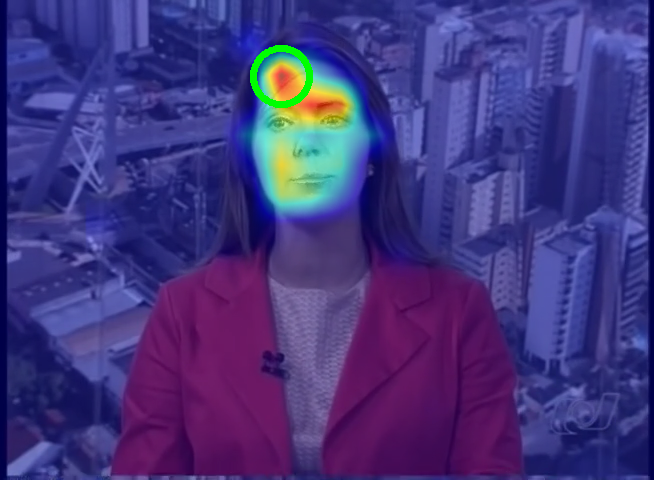} & \myimg{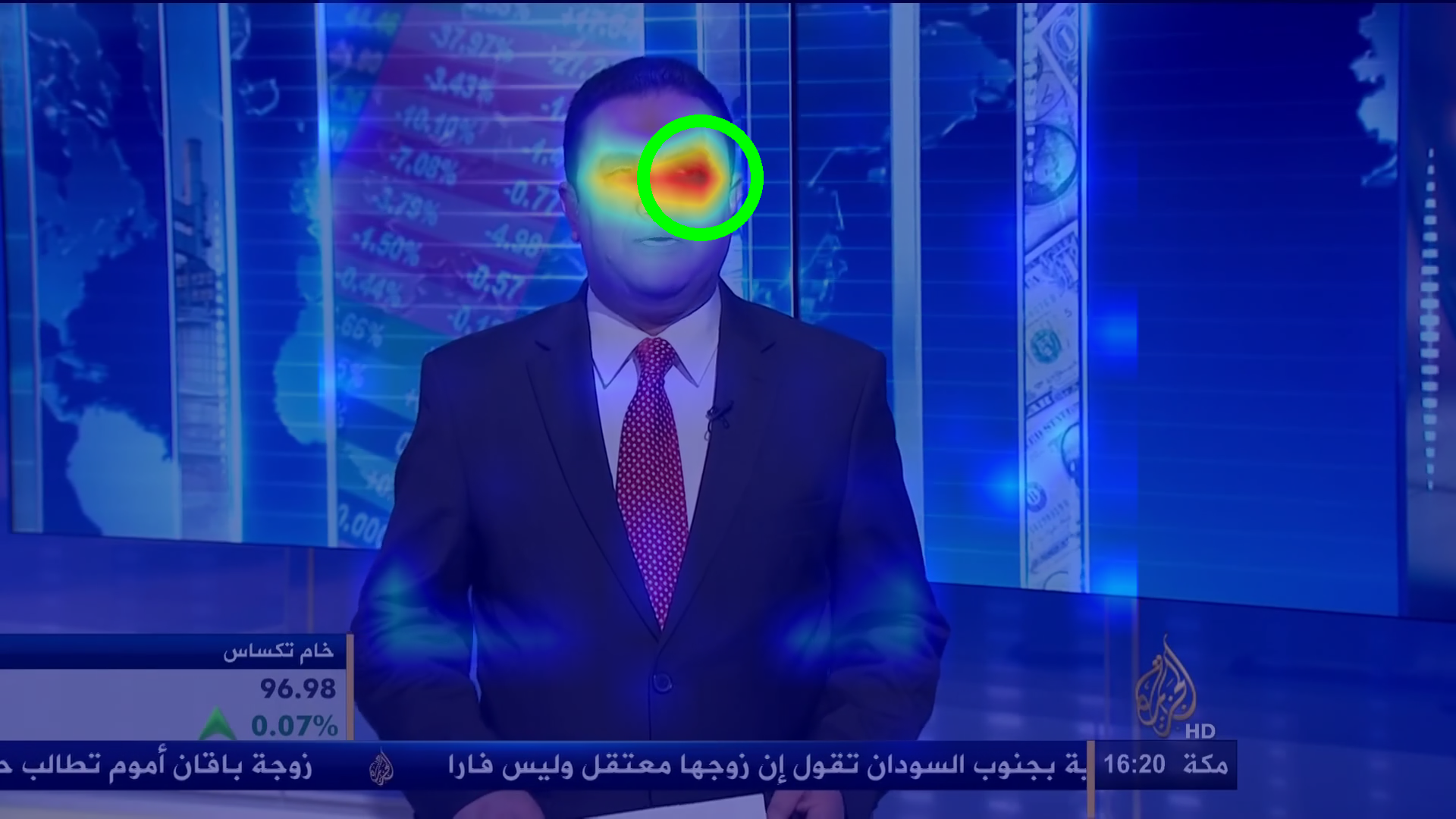} & \myimg{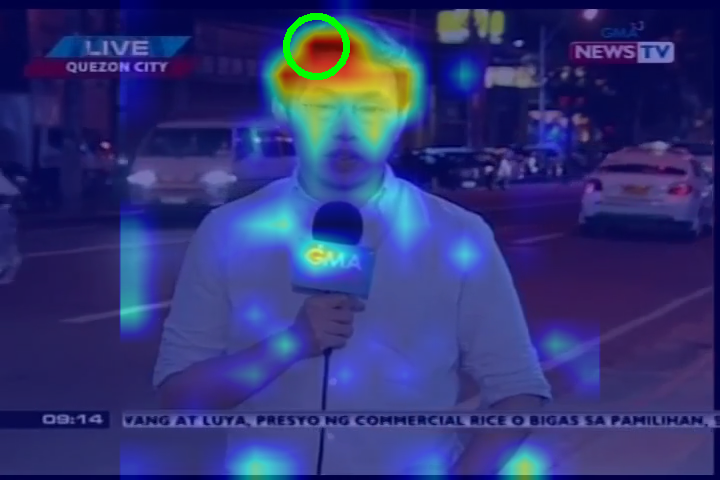} & \myimg{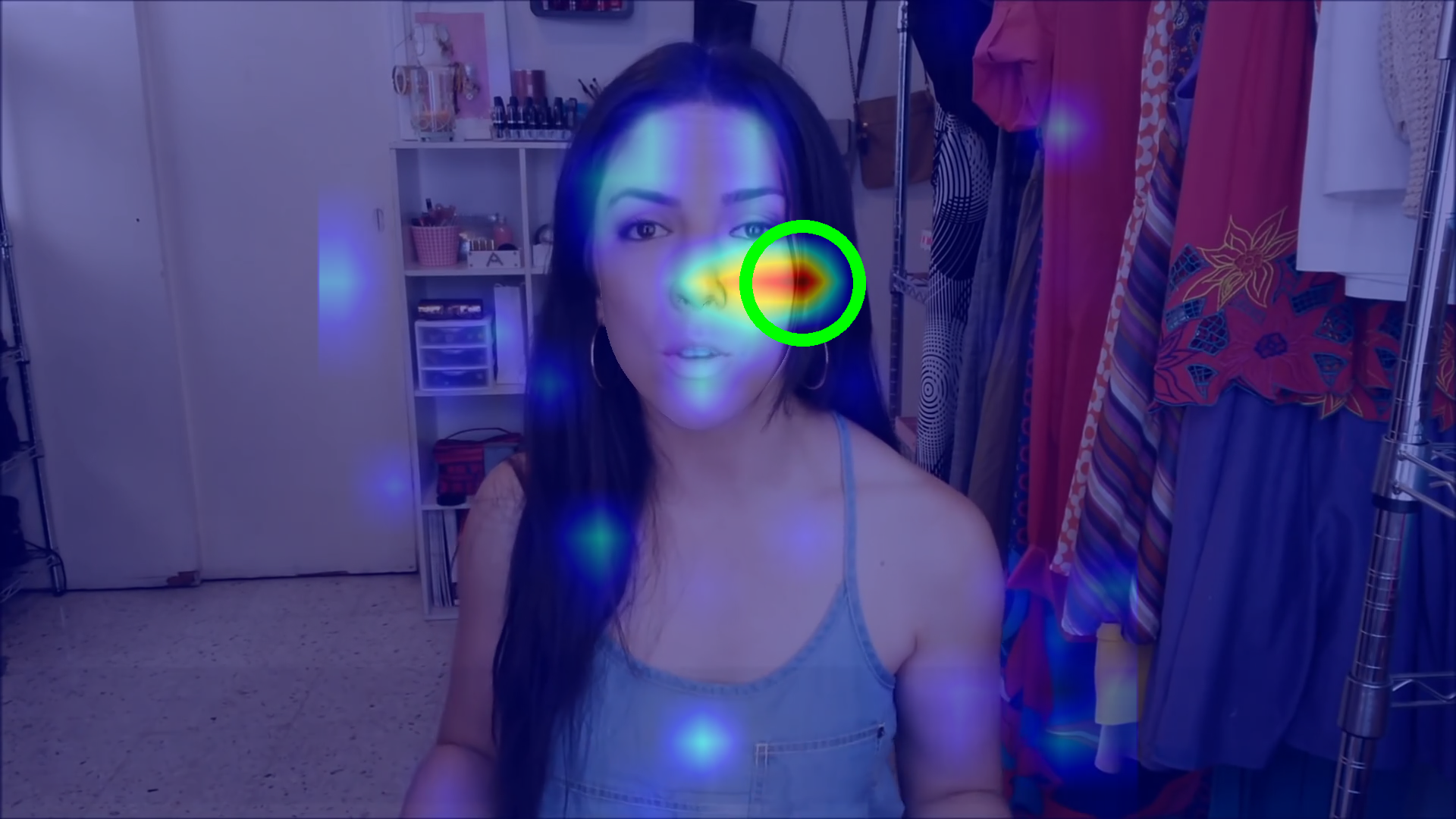} & \myimg{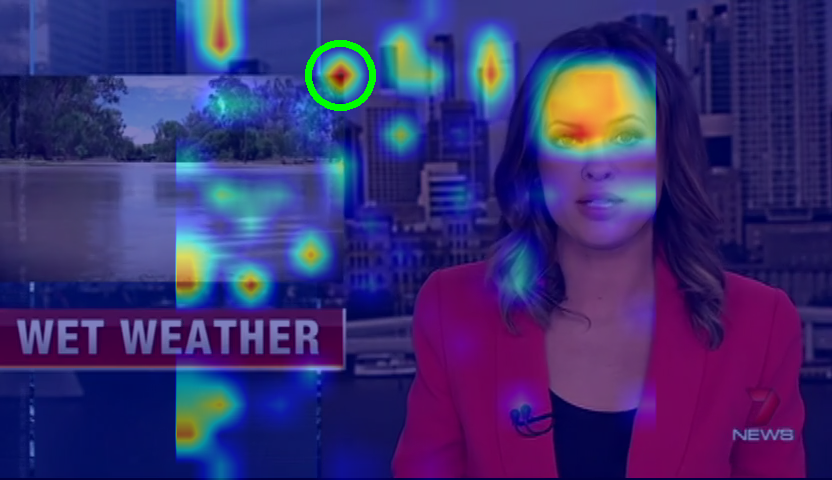} & \myimg{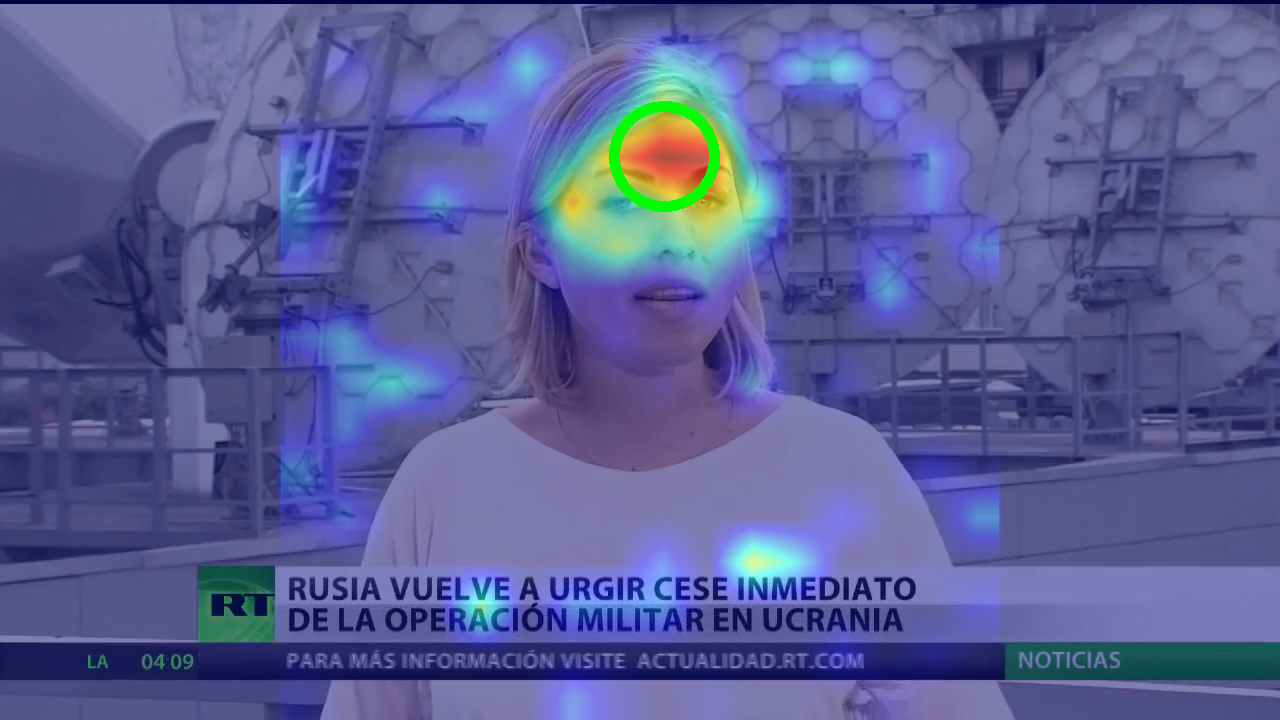}
        \end{tabular}
    \end{minipage}
    \caption{
        Alignment of spatial explanations to human annotations.
        \textit{Left:} Alignment error in terms of mean absolute error (MAE) as a function of the model confidence (fakeness score).
        The explanations align better to human annotations as the model is more confident in its predictions.
        \textit{Right:} Qualitative samples human annotation shown as the center of the red circle on top frame,
        and explanation of the CLIP-based model shown on bottom frame (maximum value indicated by the green circle).
    }
    \label{fig:spatial-explanations}
\end{figure*}

%% file: sec/8-supplementary-material.tex
\appendix

\section{Implementation details}
\label{sec:app-implementation-details}

\mypar{Linear probing.}
We train the linear probing model for $100$ epochs with early stopping (training is stopped if the validation loss does not improve for 10 consecutive epochs).
We use the Adam optimizer with a learning rate of $10^{-3}$.
For AV1M, we select a training set of 22972 videos and a validation set of 2527 videos,
both sampled from the official training set.
For FAVC, we split the entire dataset into 63$\%$ training, 7$\%$ validation, and 30$\%$ test sets and used only the real video real audio (RVRA) and fake video fake audio (FVFA) samples.

\mypar{Next-token prediction.}
The training setup for the next-token prediction is similar to the one used for the linear probing experiments.
The main difference is that we now anneal the learning rate using a cosine scheduler.
The model is trained on 50k real videos randomly sampled from the AV1M training set,
using  45k samples for training and 5k for validation.
The training set includes the real samples used for training the linear probes.

\mypar{Audio-video synchronization.}
For the audio-video synchronization task, we use the publicly released implementation of \cite{Smeu_2025_CVPR} with its default settings.
We use a temporal neighborhood of 30 frames and a learning rate scheduler with a patience of 5 epochs, a reduction factor of $0.1$, and a starting learning rate of $10^{-5}$.
The training set is identical to the one used for the next-token prediction task.

\section{Model checkpoints}
\label{sec:model-checkpoints}

For AV-HuBERT we use the \texttt{self\_large\_vox\_433h} checkpoint which was pretrained on LRS3 and VoxCeleb2 and finetuned on 433 hours of LRS3 samples for the task of visual speech recognition.
For CLIP we use the \texttt{openai/clip-vit-large-patch14} checkpoint, and for Video MAE the \texttt{MCG-NJU/videomae-large}, both which can be found on HuggingFace.
For FSFM we use the \texttt{FF++\_c23\_32frames} checkpoint, which was trained on the FaceForensics++ dataset.
For Wav2Vec2, we use the \texttt{facebook/wav2vec2-xls-r-2b} HuggingFace checkpoint of the 2-billion parameters model, which was pretrained on 436k hours of multilingual data.
For Auto-AVSR, we use the best models trained on 3,448 hours:
\texttt{LRS3\_V\_WER19.1} for visual features,
\texttt{LRS3\_A\_WER1.0} for audio features,
\texttt{LRS3\_AV\_WER0.9} for multimodal features.
For BRAVEn we use the checkpoints from the high-resource setup, with self-training, finetuned for ASR and VSR, respectively.
Input face crops and audio files are extracted using the AV-HuBERT code.

\section{Datasets}
\label{sec:app-dataset}

\mypar{General preprocessing.}
We follow the same preprocessing steps used for each backbone network during its pretraining stage.
Visual cropping is applied according to the ``pretraining content'' column in \cref{tab:models}.
Specifically, models pretrained on generic visual content use uncropped frames, those pretrained on faces use face-cropped frames, and those pretrained on lips use lip-cropped frames.

\mypar{DeepfakeEval 2024 preprocessing.}
We use TalkNet-ASD \cite{tao2021someone} to identify which segments in an audio-video media file contain a single person speaking.
We selected audio-video segments that met these criteria:
(i) each video segment that has an associated audio stream;
(ii) a video segment should contain a single speaking face that was tracked in every frame. (Some videos had static images or background music instead of speech and these were discarded);
(iii) the identified face is larger than 100px$\times$100px; 
(iv) the duration of audio-video segment is between 3 and 60 seconds.

\section{Baselines}
\label{app:baselines}

\mypar{Randomly initialized models}.
For the randomly initialized AV-HuBERT models, we keep the same architecture as the pre-trained version.
We also adopt the same initialization scheme used when training AV-HuBERT from scratch:
BatchNorm and LayerNorm layers start from constant parameters (weights set to 1 and biases to 0),
while linear and embedding layers follow a BERT-style initialization, with weights sampled from a normal distribution with mean 0 and standard deviation 0.02.
Finally, we preserve the same preprocessing pipeline as in the pre-trained setup, using log filterbanks for audio and mouth crops for the visual stream.

\mypar{AVFF} \citep{oorloff2024avff} is a multimodal, two stage deepfake detector.
In the first self-supervised stage, the model encodes both the input audio and video with two separate encoders, masks the tokens in a complementary way, predicts the masked ones using the remaining, visible tokens and then reconstructs the original input.
This stage helps the model extract meaningful information from both streams and better align the features.
In the second stage, the features extracted are used as input for a classifier network which is trained on the task of deepfake detection. For our experiments we finetuned the checkpoint pretrained on Kinetics400, available in the unofficial open version.\footnote{\url{https://github.com/JoeLeelyf/OpenAVFF}}

\mypar{SpeechForensics} \citep{liang24speechforensics} is an unsupervised method that detects deepfakes by measuring the alignment between audio and video streams. 
The alignment is computed as the cosine similarity between audio and visual features extracted from a pretrained AV-HuBERT.
To account for desynchronizations, which are common in in-the-wild real videos, the authors propose measuring the best alignment score by shifting the streams within a fixed window.
For a fair comparison, we use the same set of AV-HuBERT features, as previously used for linear probing.
Since audio and visual features sometimes have different lengths, we align them by trimming at the end.
We note, however, that the original implementation uses uniform sampling, which yields slightly better performance.
In~\cref{sec:speechforensics}, we conduct an ablation study on the pooling operation and window size parameters.

\mypar{AuViRe} \citep{koutlis2025auvire} is a supervised method trained on the task of temporal forgery localization. The model uses rich self-supervised representations (AV-HuBERT features) as input for a 1D CNN. Firstly, the model reconstructs the input features both within and cross modality (for cross modality, only the video features are reconstructed based on audio ones). Then, a classification head is used to detect the per frame manipulations and a regression head to predict the manipulation segment boundaries. The method was trained on two datasets, LAV-DF and AV-Deepfake 1M. In our experiments, we used the checkpoint trained on AV-Deepfake 1M from the official GitHub repository, together with the default parameters.

\mypar{AVAD} \citep{feng23cvpr} is an unsupervised method which focuses on modeling the distribution over delays in real data and detect fakes at test time based on the deviation from norm. This is achieved by training an autoregressor on the distribution over delays predicted by another component: the audio-visual synchronization model. The authors trained these components only on real data. For our experiments, we used the default parameters and checkpoint specified on the official GitHub repository. The checkpoint used was trained on LRS2 and LRS3 datasets.

\mypar{RealForensics} \citep{haliassos22realforensics} is a supervised two-step method that leverages self-supervised pretraining to improve robustness. In the first step, the objective is to learn temporally dense video representations using a cross-modal student-teacher framework, where each student predicts the output of the other modality's teacher. The backbones for each modalities consist of convolutional networks with one-block transformer encoder predictors on top. In the second stage, the detector is tasked with predicting the video targets produced by the video teacher from stage 1 for real videos, combined with a cross-entropy minimization loss on both real and fake  inputs. This multi-task learning encourages the model to focus on high-level facial dynamics. For our experiments, we use the model trained on FaceForensics++ \citep{rossler2019faceforensics}, with additional real samples from LRW \citep{Chung2016LipRI}.

\section{Further results}
\label{sec:app-further-results}

\subsection{Average precision results}
\label{app:average-precision}

\input{tables/tab-main-ap}

\input{tables/tab-proxy-tasks-ap}

We complement the results from~\cref{tab:main-results-v2} and~\cref{tab:ntp-sync-sup-results-v2} with average precision (AP) scores reported in~\cref{tab:main-results_AP} and~\cref{tab:ntp-sync-sup-results_AP}, respectively.

\subsection{Synchronization of pretrained representations}
\label{sec:speechforensics}

If audio-visual self-supervised features are trained jointly, then they are already aligned to a degree.
Here, we investigate how well this pre-existing alignment works for the task of deepfake detection.
This contrasts with the additional alignment introduced in \cref{subsec:proxy-tasks}, which was required when combining representations from different models.

We use audio-only and visual-only features from AV-HuBERT and measure their alignment via cosine similarity.
The fakeness score $s$ between the audio features $\rva$ and visual features $\rvv$ is defined as a generalization of \cite{reiss2023detecting,liang24speechforensics}:
\begin{equation}
s(\rva, \rvv) = \min_{\delta\in[-\Delta, +\Delta]} \mathrm{pool}_t \left( - \cos(\rva_t, \rvv_{t+\delta})\right),
\label{eq:sync-pretrained}
\end{equation}
where
\begin{itemize}
    \item $\delta$ is a temporal shift that compensates for imperfect synchronization of real video \cite{feng23cvpr}.
    Allowing moderate shifts can therefore improve discrimination between real and fake samples.
    We test both strict ($\Delta = 0$) and moderate alignment ($\Delta = 15$), following \cite{liang24speechforensics}.
    \item $\mathrm{pool}$ aggregates per-frame fakeness scores.
    We experiment with several pooling strategies: average, min, max, 3rd and 97th percentiles, and a scaled log-sum-exp (with temperature given by the length of the sequence).
\end{itemize}

\cref{eq:sync-pretrained} subsumes prior methods as special cases:
FACTOR \cite{reiss2023detecting} corresponds to $\Delta=0$ and 97th-percentile pooling;
SpeechForensics \cite{liang24speechforensics} corresponds to $\Delta=15$ with average pooling.

\cref{tab:ablation_speech} shows results on the four datasets introduced in \cref{sec:datasets}.
At a high level, all variants behave similarly when compared to a supervised baseline (linear probing on AV-HuBERT multimodal features):
they underperform on AV1M and perform better on AVLips and DFE-2024.
If we look closer, and at the two axes, we first see that average pooling, as used in SpeechForensics \cite{liang24speechforensics}, is rarely optimal.
This is especially evident for AV1M, where the max-based pooling variants perform best.
This behavior aligns with the nature of the dataset: since AV1M contains local manipulations, max pooling is able to capture these brief artifacts.
Interestingly, for AVLips and DFE-2024, it is the minimum pooling that yields consistent (albeit small) gains.
Compared to max pooling, which searches for ``evidence of fakeness somewhere,'' min pooling searches for ``evidence of realness anywhere.''
This is particularly suitable for real in-the-wild videos, whose synchronization is imperfect and which contain only a few strongly aligned segments.
In terms of feature alignment by shifting, we see that it does help in certain setups, most notably for DFE-2024.

\begin{figure}
    \centering
    \includegraphics[width=1\linewidth]{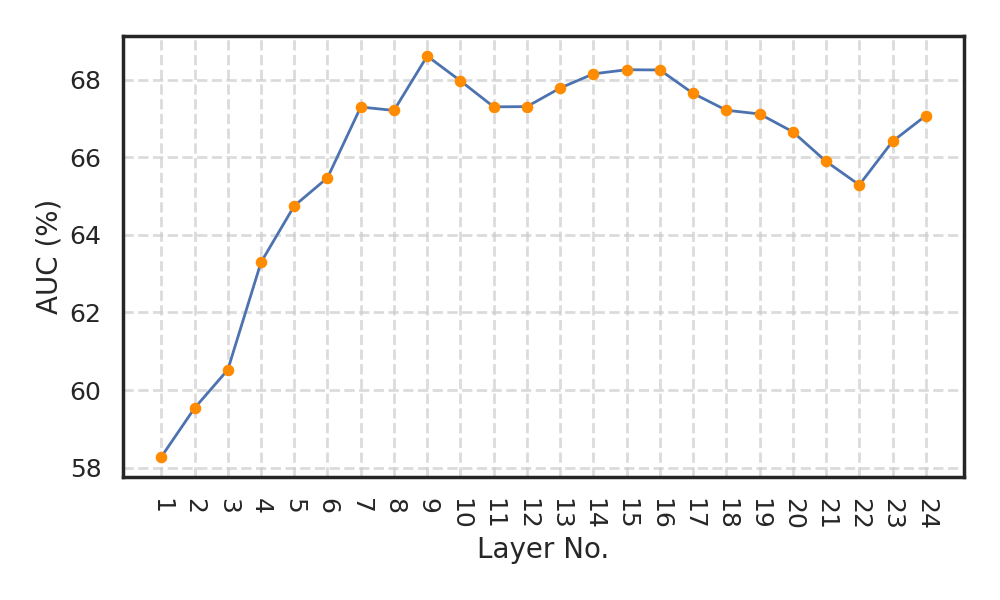}
    \caption{
        Performance on the AV1M dataset using the pretrained AV-HuBERT representations extracted from different layers.
        We use the SpeechForensics approach that directly compares AV-HuBERT features over a optimal shifting window.
    }
    \label{fig:sf-on-layer}
\end{figure}

\textbf{Layer-wise analysis.}
We further analyze the representational capabilities of AV-HuBERT features across all 24 transformer layers.
This analysis is conducted on the AV1M test set and, %
using SpeechForensics \cite{liang24speechforensics} with its default hyperparameters (average pooling and a shift window of 15).
As shown in \cref{fig:sf-on-layer}, performance roughly stabilizes from  layer 7, with a small drop at layer 22. 
Interestingly, the best performance is obtained at layer 9 (68.6\%), but the improvement over the last layer is not substantial.
Note that the last-layer performance in \cref{fig:sf-on-layer} (67.1\%) is slightly lower than the corresponding value in \cref{tab:ablation_speech} (68.1\%; for average pooing and $\Delta = 15$ on AV1M);
this discrepancy is due to the absence of LayerNorm in the layer-wise analysis.

\begin{table*}
    \centering
    \small
    \begin{tabularx}{\linewidth}{X rr rr rr rr} %
        \toprule
        & \multicolumn{2}{c}{FAVC} 
        & \multicolumn{2}{c}{AV1M} 
        & \multicolumn{2}{c}{AVLips} 
        & \multicolumn{2}{c}{DFE-2024} \\
        \cmidrule(lr){2-3}\cmidrule(lr){4-5}\cmidrule(lr){6-7}\cmidrule(lr){8-9}
              Pooling function
              & $\Delta=0$ & $\Delta=15$ 
              & $\Delta=0$ & $\Delta=15$ 
              & $\Delta=0$ & $\Delta=15$ 
              & $\Delta=0$ & $\Delta=15$
              \\
\midrule
average       & 99.3     & \bf 100 & 65.3     & 68.1     & 94.8     & 92.7     & 62.7     & 75.6 \\
\midrule                            
\multicolumn{9}{l}{\it Maximum variants} \\
max           & 99.3     & 99.1    & \bf 76.8 & 76.9     & 94.8     & 93.7     & 61.0     & 64.9 \\
log-sum-exp   & 99.4     & 99.6    & \bf 76.8 & 76.9     & 95.3     & 94.2     & 60.8     & 64.4 \\
percentile-97 & \bf 99.5 & 99.4    & 75.0     & \bf 80.0 & \bf 95.4 & 93.0     & 61.4     & 68.8 \\
\midrule                            
\multicolumn{9}{l}{\it Minimum variants} \\
min           & 98.2     & 99.3    & 58.1     & 56.1     & 94.0     & 96.1     & 66.3     & 74.4 \\
percentile-3  & 99.1     & 99.5    & 59.9     & 58.6     & \bf 95.4 & \bf 96.4 & \bf 63.9 & \bf 75.7 \\
\midrule
\multicolumn{9}{l}{\it Supervised baseline} \\
AV-H $+$ linear on FAVC (row \mylabel{14} \cref{tab:main-results-v2}) &
\multicolumn{2}{c}{\color{gray} 100} &
\multicolumn{2}{c}{94.5} &
\multicolumn{2}{c}{78.5} &
\multicolumn{2}{c}{58.2} \\
    \bottomrule
    \end{tabularx}
    \caption{
        AUC performance (\%) when directly measuring the alignment of pretrained AV-HuBERT audio and visual features. 
        We evaluate over pooling functions and temporal shifts ($\Delta$), to compensate desynchronizations of real in-the-wild videos.
        Bold indicates best results in each column except for the supervised baseline (multimodal AV-HuBERT features with a linear classifier trained on the FAVC dataset).
    }
    \label{tab:ablation_speech}
\end{table*}

\subsection{Classification head analysis}
\label{app:classification-head-analysis}

We compare the linear classifier described in \cref{sec:methodology} with a more powerful head: a transformer.
The transformer has width 768, 4 layers, 8 attention heads. Representations are projected using a linear layer to the input 768 dimension.
To obtain a classification prediction,
we use the [CLS] token and project it through a linear layer.
The results are displayed in \cref{tab:linear_vs_transformer}.
We observe that, in most cases, the performance obtained by the transformer model is on par with or slightly below that of linear probing.
The single notable exception observed occurs on FSFM features when trained on AV1M and tested on FAVC.
In this scenario, performance increases significantly from $40.9\%$ AUC to $72.3\%$ AUC.

\input{tables/tab-classifier-head}

\subsection{Combinations of features}
\label{app:combinations-of-features}

\begin{figure}
    \centering
    \includegraphics[width=1\linewidth]{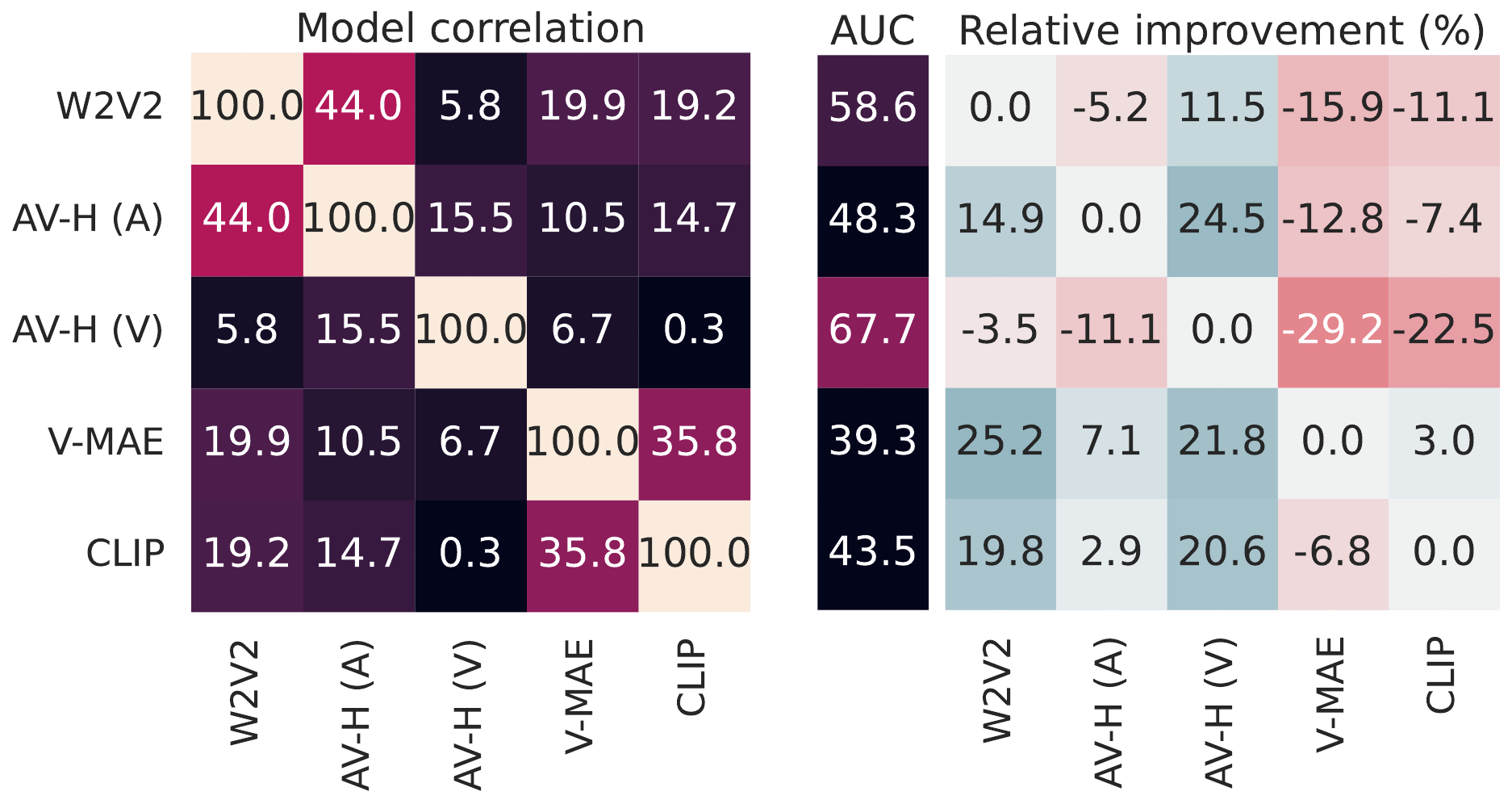}
    \caption{
        Correlations between models (left) and downstream performance (right).
        Training was done on AV1M, testing on DFEval-2024.
    }
    \label{fig:model-comb-av1m-dfeval}
\end{figure}

In~\cref{fig:model-comb-av1m-dfeval} we show the results for combinations of features when testing is done on DFE-2024.
Compared to~\cref{fig:model-combination} (testing done on FAVC), certain trends are much better highlighted:
first, the performance of every model greatly increases when combined with Wav2Vec2 or AV-HuBERT (V);
second, the correlations between models' results are weaker, suggesting that the models capture more distinct patterns.

%% file: tables/tab-main-ap.tex
\begin{table*}
    \definecolor{mycolor}{HTML}{d4a373}
    \newcolumntype{i}{r}
    \newcolumntype{o}{r}
    \centering
    \small
    \tabcolsep 4.5pt
    \begin{tabularx}{\linewidth}{r X c io io oo ioo o}
        \toprule
        &
        & 
        & \multicolumn{1}{c}{\ii{A}}
        & \multicolumn{1}{c}{\ii{B}}
        & \multicolumn{1}{c}{\ii{C}}
        & \multicolumn{1}{c}{\ii{D}}
        & \multicolumn{1}{c}{\ii{E}}
        & \multicolumn{1}{c}{\ii{F}}
        & \multicolumn{1}{c}{\ii{G}}
        & \multicolumn{1}{c}{\ii{H}}
        & \multicolumn{1}{c}{\ii{I}}
        & \multicolumn{1}{c}{\ii{J}}
        \\
        &
        &
        & \multicolumn{2}{c}{Test on FAVC}
        & \multicolumn{2}{c}{Test on AV1M}
        & \multicolumn{2}{c}{Test on AVLips}
        & \multicolumn{3}{c}{Test on DFE-2024}
        \\
        \cmidrule(lr){4-5}
        \cmidrule(lr){6-7}
        \cmidrule(lr){8-9}
        \cmidrule(lr){10-12}
        &
        &
        & \multicolumn{1}{c}{FAVC} & \multicolumn{1}{c}{AV1M}
        & \multicolumn{1}{c}{AV1M} & \multicolumn{1}{c}{FAVC}
        & \multicolumn{1}{c}{FAVC} & \multicolumn{1}{c}{AV1M}
        & \multicolumn{1}{c}{DFE} & \multicolumn{1}{c}{FAVC} & \multicolumn{1}{c}{AV1M}
        \\
        &
        &
        & \multicolumn{1}{c}{$\downarrow$}
        & \multicolumn{1}{c}{$\downarrow$}
        & \multicolumn{1}{c}{$\downarrow$}
        & \multicolumn{1}{c}{$\downarrow$}
        & \multicolumn{1}{c}{$\downarrow$}
        & \multicolumn{1}{c}{$\downarrow$}
        & \multicolumn{1}{c}{$\downarrow$}
        & \multicolumn{1}{c}{$\downarrow$}
        & \multicolumn{1}{c}{$\downarrow$} 
        & \multicolumn{1}{c}{mean}
        \\
        & Model            
        & Modality
        & \multicolumn{1}{c}{FAVC} & \multicolumn{1}{c}{FAVC}
        & \multicolumn{1}{c}{AV1M} & \multicolumn{1}{c}{AV1M}
        & \multicolumn{1}{c}{AVLips} & \multicolumn{1}{c}{AVLips}
        & \multicolumn{1}{c}{DFE} & \multicolumn{1}{c}{DFE} & \multicolumn{1}{c}{DFE}& \multicolumn{1}{c}{OOD}
        \\
        \midrule
        \ii{1}   & AV-HuBERT (A) random & audio        & 100     & 99.9     & 99.1     & 91.1     & \gray 55.7     & \gray 60.2     & \gray 89.2     & \gray 85.2     & \gray 87.1     & 79.9 \\
        \ii{2}   & AV-HuBERT (V) random & visual       & 99.0    & 96.2     & 52.3     & 51.3     & \gray 63.2     & \gray 59.5     & \gray 88.4     & \gray 91.6     & \gray 88.8     & 75.1 \\
        \midrule
        \ii{3}   & AV-HuBERT (A)        & audio        & \bf 100 & \bf 100  & \bf 100  & \bf 99.2 & \gray   55.6   & \gray 62.9     & \gray 93.4     & \gray  89.1    & \gray  88.9    & 82.6 \\
        \ii{4}   & Auto-AVSR (ASR)      & audio        & 100     & 98.4     & 96.8     & 51.2     & \gray 58.3     & \gray 55.3     & \gray 93.4     & \gray 88.6     & \gray 88.8     & 73.4 \\
        \ii{5}   & Wav2Vec2             & audio        & \bf 100 & \bf 100  & \bf 100  & 96.5     & \gray \bf 61.9 & \gray \bf 64.3 & \gray 92.5     & \gray \bf 92.5 & \gray \bf 91.7 & \bf 84.5 \\
        \ii{6}   & BRAVEn (A)           & audio        & \bf 100 & \bf 100  & 99.9     & 88.7     & \gray 51.2     & \gray 53.9     & \gray \bf 94.0 & \gray 92.4     & \gray 90.4     & 79.4 \\
        \midrule
        \ii{7}   & AV-HuBERT (V)        & visual       & \bf 100 & 99.6     & 94.7     & 62.1     & 98.4           & 88.6           & 95.1           & 92.9           & 92.8           & 89.1 \\
        \ii{8}   & Auto-AVSR (VSR)      & visual       & 99.9    & 98.7     & 60.3     & 52.6     & 86.2           & 74.3           & 92.9           & 88.8           & 90.3           & 81.8 \\
        \ii{9}   & FSFM                 & visual       & 99.9    & 95.2     & 95.5     & 53.6     & 83.6           & 47.1           & 93.8           & \bf 94.5       & 84.3           & 76.4 \\
        \ii{10}  & CLIP VIT-L/14        & visual       & \bf 100 & 99.8     & 96.8     & \bf 71.5 & 65.3           & 60.9           & 93.8           & 90.7           & 85.2           & 78.9 \\
        \ii{11}  & Video-MAE-large      & visual       & \bf 100 & 98.1     & \bf 99.7 & 61.3     & 73.9           & 53.2           & 89.3           & 86.9           & 84.3           & 76.3 \\
        \ii{12}  & BRAVEn (V)           & visual       & \bf 100 & \bf 99.9 & 94.1     & 63.7     & \bf 98.8       & \bf 96.9       & \bf 95.3       & 93.2           & \bf 95.5       & \bf 91.3 \\
        \midrule
        \ii{13}  & AV-HuBERT            & audio-visual & \bf 100 & \bf 100  & \bf 99.9 & \bf 94.9 & \bf 76.5       & \bf 82.1       & \bf 94.7       & \bf 90.8       & \bf 88.7       & \bf 88.8 \\
        \ii{14}  & Auto-AVSR            & audio-visual & 99.7    & 97.7     & 92.8     & 54.4     & 63.7           & 58.4           & 92.7           & 85.9           & 89.1           & 74.9 \\
        \bottomrule
    \end{tabularx}
    \caption{%
        Average precision (AP, \%) performance of linear probes trained on multiple self-supervised representations.
    }
    \label{tab:main-results_AP}
\end{table*}

%% file: tables/tab-proxy-tasks-ap.tex
\begin{table}
    \centering
    \setlength{\tabcolsep}{3pt}
    \footnotesize
    \begin{tabularx}{\linewidth}{X rrr rrr}
        \toprule
                            & \multicolumn{3}{c}{AV1M} & \multicolumn{3}{c}{FAVC} \\
                            \cmidrule(lr){2-4}
                            \cmidrule(lr){5-7}
        Model               & Sup. & NTP       & Sync.     & Sup. & NTP       & Sync. \\
        \midrule
        \multicolumn{5}{l}{\textit{Single features}} \\
        AV-HuBERT (A)       & \bf 99.2 & \bf 91.8 & \na & \bf 100 & \bf 98.8 & \na \\
        Wav2Vec2            & 96.5 & 56.4 & \na & \bf 100 & 96.6 & \na \\
        AV-HuBERT (V)       & 62.1 & 49.2 & \na & 99.6 & 97.1 & \na \\
        CLIP                & 71.5 & 49.7 & \na & 99.8 & 96.8 & \na \\
        \midrule
        \multicolumn{5}{l}{\textit{Combination of features}} \\
        AV-H (A + V) (rand.)  & 78.3 & 64.4 & 50.8 & 99.8 & 98.1 & 96.0 \\
        AV-H (A + V)          & 97.2 & 83.4 & \bf 85.7  & \bf 100 & \bf 99.5 & \bf 99.7 \\
        AV-H (A) + CLIP       & \bf 99.2 & \bf 88.0 & 50.0 & \bf 100 & 98.6 & 96.1  \\
        W2V2 + AV-H (V)       & 95.7 & 59.4 & 85.1 & \bf 100 & 98.6 & 99.6 \\
        W2V2 + CLIP           & 97.0 & 56.6 & 50.5 & \bf 100 & 97.6 & 90.7 \\
        \bottomrule
    \end{tabularx}
    \caption{%
        Average precision (AP, \%) performance for deepfake detection when training for the two anomaly detection proxy tasks: next-token prediction (NTP) and audio-video synchronization (sync.). 
        Supervised models (sup.) are trained cross-domain (FAVC$\to$AV1M and AV1M$\to$FACV, respectively).
        Anomaly detection models are trained on real data only (a subset of VoxCeleb).
    }
    \label{tab:ntp-sync-sup-results_AP}
\end{table}

%% file: tables/tab-classifier-head.tex
\begin{table*}
    \definecolor{mycolor}{HTML}{d4a373}
    \definecolor{tblue}{HTML}{2E5A88}
    \newcolumntype{i}{r}
    \newcolumntype{o}{r}
    \newcommand{\trf}{\color{tblue}}
    \centering
    \small
    \begin{tabularx}{\linewidth}{r X c io io oo ioo o}
        \toprule
        & & & \multicolumn{1}{c}{\ii{A}} & \multicolumn{1}{c}{\ii{B}} & \multicolumn{1}{c}{\ii{C}} & \multicolumn{1}{c}{\ii{D}} & \multicolumn{1}{c}{\ii{E}} & \multicolumn{1}{c}{\ii{F}} & \multicolumn{1}{c}{\ii{G}} & \multicolumn{1}{c}{\ii{H}} & \multicolumn{1}{c}{\ii{I}} & \multicolumn{1}{c}{\ii{J}} \\
        & & & \multicolumn{2}{c}{Test on FAVC} & \multicolumn{2}{c}{Test on AV1M} & \multicolumn{2}{c}{Test on AVLips} & \multicolumn{3}{c}{Test on DFE-2024} & \\
        \cmidrule(lr){4-5} \cmidrule(lr){6-7} \cmidrule(lr){8-9} \cmidrule(lr){10-12}
        & & & \multicolumn{1}{c}{FAVC} & \multicolumn{1}{c}{AV1M} & \multicolumn{1}{c}{AV1M} & \multicolumn{1}{c}{FAVC} & \multicolumn{1}{c}{FAVC} & \multicolumn{1}{c}{AV1M} & \multicolumn{1}{c}{DFE} & \multicolumn{1}{c}{FAVC} & \multicolumn{1}{c}{AV1M} & \\
        & & & \multicolumn{1}{c}{$\downarrow$} & \multicolumn{1}{c}{$\downarrow$} & \multicolumn{1}{c}{$\downarrow$} & \multicolumn{1}{c}{$\downarrow$} & \multicolumn{1}{c}{$\downarrow$} & \multicolumn{1}{c}{$\downarrow$} & \multicolumn{1}{c}{$\downarrow$} & \multicolumn{1}{c}{$\downarrow$} & \multicolumn{1}{c}{$\downarrow$} & \multicolumn{1}{c}{mean} \\
        & Model & Head & \multicolumn{1}{c}{FAVC} & \multicolumn{1}{c}{FAVC} & \multicolumn{1}{c}{AV1M} & \multicolumn{1}{c}{AV1M} & \multicolumn{1}{c}{AVLips} & \multicolumn{1}{c}{AVLips} & \multicolumn{1}{c}{DFE} & \multicolumn{1}{c}{DFE} & \multicolumn{1}{c}{DFE} & \multicolumn{1}{c}{OOD} \\
        \midrule
                                 & \multicolumn{12}{l}{\textit{Audio features}} \\
        \multirow{2}{*}{\ii{3}}  & \multirow{2}{*}{AV-HuBERT (A)}                         & L      & 100.0      & 100.0     & 100.0      & 99.0      & \gray 50.0 & \gray 57.2 & \gray 65.8 & \gray 49.1 & \gray 48.3 & 67.3      \\
                                 &                                                        & \trf T & \trf 100.0 & \trf 99.9 & \trf 99.9  & \trf 88.8 & \gray 47.7 & \gray 53.4 & \gray 70.2 & \gray 48.2 & \gray 55.0 & \trf 65.5 \\
        \cmidrule(lr){2-13}                                                                        
        \multirow{2}{*}{\ii{4}}  & \multirow{2}{*}{Auto-AVSR (ASR)}                       & L      & 99.7       & 76.0      & 96.4       & 50.3      & \gray 52.9 & \gray 49.6 & \gray 63.5 & \gray 49.4 & \gray 47.5 & 54.3      \\
                                 &                                                        & \trf T & \trf 100.0 & \trf 80.1 & \trf 97.3  & \trf 51.3 & \gray 46.4 & \gray 49.4 & \gray 70.3 & \gray 51.6 & \gray 50.4 & \trf 54.9 \\
        \cmidrule(lr){2-13}                                                                        
        \multirow{2}{*}{\ii{5}}  & \multirow{2}{*}{Wav2Vec2}                              & L      & 100.0      & 99.9      & 100.0      & 96.6      & \gray 51.3 & \gray 56.3 & \gray 58.7 & \gray 62.3 & \gray 58.6 & 70.8      \\
                                 &                                                        & \trf T & \trf 100.0 & \trf 99.9 & \trf 100.0 & \trf 94.2 & \gray 47.3 & \gray 56.5 & \gray 50.6 & \gray 52.5 & \gray 58.2 & \trf 68.1 \\
        \midrule                                                                                   
                                 & \multicolumn{12}{l}{\textit{Visual features}} \\                
        \multirow{2}{*}{\ii{6}}  & \multirow{2}{*}{AV-HuBERT (V)}                         & L      & 100.0      & 95.5      & 93.7       &  64.1 & 98.3      & 90.5       & 72.1       & 63.7       & 67.7      & 80.0      \\
                                 &                                                        & \trf T & \trf 100.0 & \trf 92.9 & \trf 93.7  & \trf 58.6 & \trf 98.6  & \trf 88.3  & \trf 73.8  & \trf 70.3  & \trf 61.8  & \trf 78.4 \\
        \cmidrule(lr){2-13}                                                                        
        \multirow{2}{*}{\ii{7}}  & \multirow{2}{*}{Auto-AVSR (VSR)}                       & L      & 97.8       & 77.5      & 59.0  &   51.3   & 83.3      & 70.1       & 64.3       & 48.7       & 56.1        & 64.5      \\
                                 &                                                        & \trf T & \trf 98.5  & \trf 75.5 & \trf 56.9  & \trf 51.4 & \trf 83.4  & \trf 63.4  & \trf 62.4  & \trf 46.9  & \trf 56.2  & \trf 62.8 \\
        \cmidrule(lr){2-13}                                                                        
        \multirow{2}{*}{\ii{8}}  & \multirow{2}{*}{FSFM}                                  & L      & 97.1       & 40.9      & 95.3       &  52.7 & 84.3      & 36.8       & 71.7       & 71.8       & 43.5       & 55.0      \\
                                 &                                                        & \trf T & \trf 98.8  & \trf 72.3 & \trf 99.0  & \trf 47.9 & \trf 79.8  & \trf 42.5  & \trf 70.7  & \trf 68.5  & \trf 48.5  & \trf 59.9 \\
        \cmidrule(lr){2-13}                                                                        
        \multirow{2}{*}{\ii{9}}  & \multirow{2}{*}{CLIP ViT-L/14}                         & L      & 99.8       & 95.2      & 96.5  & 71.1     & 60.3      & 53.3       & 73.9       & 55.6       & 43.5        & 63.2      \\
                                 &                                                        & \trf T & \trf 99.7  & \trf 96.4 & \trf 99.2  & \trf 62.3 & \trf 57.0  & \trf 59.1  & \trf  73.3 & \trf 55.8  & \trf 50.2  & \trf 63.4 \\
        \cmidrule(lr){2-13}                                                                        
        \multirow{2}{*}{\ii{10}} & \multirow{2}{*}{Video-MAE-large}                       & L      & 100.0      & 70.4      & 99.8     & 60.0  & 71.3      & 47.2       & 54.5       & 45.6       & 39.3          & 55.6      \\
                                 &                                                        & \trf T & \trf 100.0 & \trf 60.8 & \trf 100.0 & \trf 53.0 & \trf 81.3  & \trf 56.5  & \trf 52.9  & \trf 58.9  & \trf 41.2  & \trf 58.6 \\
        \midrule                                                                                   
                                 & \multicolumn{12}{l}{\textit{Audio-visual features}} \\          
        \multirow{2}{*}{\ii{11}} & \multirow{2}{*}{AV-HuBERT}                             & L      & 100.0      & 99.5      & 99.9   & 94.5     & 78.5      & 84.4       & 70.4       & 58.2       & 54.3          & 78.2      \\
                                 &                                                        & \trf T & \trf 100.0 & \trf 99.6 & \trf 99.9  & \trf 78.6 & \trf 82.5  & \trf 76.8  & \trf 68.9  & \trf 59.4  & \trf  55.9  & \trf 75.5 \\
        \cmidrule(lr){2-13}                                                                        
        \multirow{2}{*}{\ii{12}} & \multirow{2}{*}{Auto-AVSR}                             & L      & 94.7       & 68.3      & 91.6   & 53.2    & 59.6      & 54.6       & 61.2       & 43.0       & 49.2       & 54.7      \\
                                 &                                                        & \trf T & \trf 99.7  & \trf 65.5 & \trf 93.4  & \trf 51.1 & \trf 51.8  & \trf 51.7  & \trf 66.5  & \trf 51.3  & \trf 51.9  & \trf 53.9 \\
        \bottomrule
    \end{tabularx}
    \caption{AUC performance (\%) when training a linear layer ($L$) vs. a transformer ($T$) (indicated in blue) classifier head on top of features.}
    \label{tab:linear_vs_transformer}
\end{table*}